\documentclass[]{fairmeta}
\title{Calibrating Verbal Uncertainty as a Linear Feature to Reduce Hallucinations}

\author[1,3\dagger*]{Ziwei Ji}
\author[1,4\dagger*]{Lei Yu}
\author[1]{Yeskendir Koishekenov}
\author[1,3\dagger]{Yejin Bang}
\author[2]{Anthony Hartshorn}
\author[2]{Alan Schelten}
\author[2]{Cheng Zhang}
\author[1,3]{Pascale Fung}
\author[1]{Nicola Cancedda}

\affiliation[1]{Meta FAIR}
\affiliation[2]{Meta GenAI}
\affiliation[3]{Hong Kong University of Science and Technology}
\affiliation[4]{University of Toronto}
\contribution[*]{Equal Contribution}
\contribution[\dagger]{Work done during Internship at Meta FAIR}

\abstract{LLMs often adopt an assertive language style also when making false claims. Such ``overconfident hallucinations'' mislead users and erode trust. 
Achieving the ability to express in language the actual degree of uncertainty around a claim is therefore of great importance.
We find that ``verbal\footnote{In this paper, we employ the term `verbal' to mean `pertaining to words rather than meaning or substance,' as opposed to `spoken rather than written' (refer to Merriam-Webster's definitions:  \url{https://www.merriam-webster.com/dictionary/verbal}). Readers may substitute `verbal uncertainty' with `expressed uncertainty' throughout the text if they find it preferable.} uncertainty'' is governed by a single linear feature in the representation space of LLMs, and show that this has only moderate correlation with the actual ``semantic uncertainty'' of the model. 
We apply this insight and show that (1) the mismatch between semantic and verbal uncertainty is a better predictor of hallucinations than semantic uncertainty alone and (2) we can intervene on verbal uncertainty at inference time and reduce confident hallucinations on short-form answers, achieving an average relative reduction of \textasciitilde 30\%.~\footnote{The code is available at~\url{https://github.com/facebookresearch/verbal_uncertainty_feature_calibration}.}
}

\date{\today}
\correspondence{zjiad@connect.ust.hk; jadeleiyu@cs.toronto.edu; \{pascalefung; ncan\}@meta.com}

\usepackage{colortbl}
\usepackage{enumitem}
\usepackage{multirow}
\usepackage{amssymb}
\usepackage{amsmath}
\usepackage{nicematrix}
\usepackage{booktabs}

\begin{document}

\maketitle

\section{Introduction}
\label{section:intro}
% Motivation: model's responses don't reflect the uncertainty though it is not certain. Thus, model often say answers confidently even if it achieved low semantic uncertainty level.

% Method: we study linguistic uncertainty -- finding way to calcualte. find the linear feature from the hidden state. VU and SU are not matching because the model response is not reflecting its uncertainty about the answer (misalignment). 

% Mitigation: By reducing this misalignment, we mitigate hallucination -- showing the uncertinaty to reduce potential harm (confident hallucination.) Here, hallucination we mitigate is not increasing the factuality or correctness of the answer.

% With the success of deployed large language models (LLMs), natural language is becoming the de facto interface for humans to interact with machine learning systems across various tasks, such as information seeking, summarization, and image captioning, to make real-world decisions. Natural language provides an opportunity for LLMs to generate responses that are not only informative but also nuanced, helping to achieve the user's goals. 

% Large Language Models (LLMs) have demonstrated remarkable capabilities in utilizing their internal knowledge. However, 
Despite their remarkable capability in utilizing their internal knowledge, LLMs often suffer from hallucinations, stating or implying facts that are not supported neither by their input nor by their training data \citep{ji2022survey, xiao2021hallucination, bang2023multitask, xiong2023can}. The issue is exacerbated when models produce hallucinations using language that suggests high confidence. Such overconfidence can cause users to rely too heavily on these responses \citep{zhou-etal-2024-relying, 10.1145/3630106.3658941}, possibly resulting in harm, loss of trust in the model, or both. 

While enhancing a model's ability to generate accurate knowledge is important, it is inevitable that models have knowledge gaps. In such cases, it is important for models to express uncertainty about their knowledge or altogether abstain from answering~\citep{tomani2024uncertainty, feng-etal-2024-dont, zhou-etal-2023-navigating, zhang-etal-2024-r}.
We refer to this expression as "verbal uncertainty (VU)" (see \S~\ref{subsec:verbal_u}).
When faced with questions close to their knowledge boundary, they should qualify their answers with expressions such as: ``I am not sure but ...'', and when the answer is squarely beyond such boundary, they should reply: ``I don't know''.
% For example, when faced with questions beyond their knowledge boundary, models should respond with "I don't know." % When the models are asked questions beyond their knowledge boundary, they should be able to say ``I don't know''. 
% An inaccurate answer introduced by ``I am not sure but...'' is much less problematic than the same answer assertively preceded by ``The answer is ...''. 
However, LLMs lack reliable mechanisms to convey their intrinsic confidence in the correctness of generated content by means of the degree of doubt expressed in their outputs \citep{zhou2024relying}.

\begin{figure}[t]
    \centering
    \begin{minipage}[t]{0.6\linewidth}
        \centering
        \includegraphics[width=\linewidth]{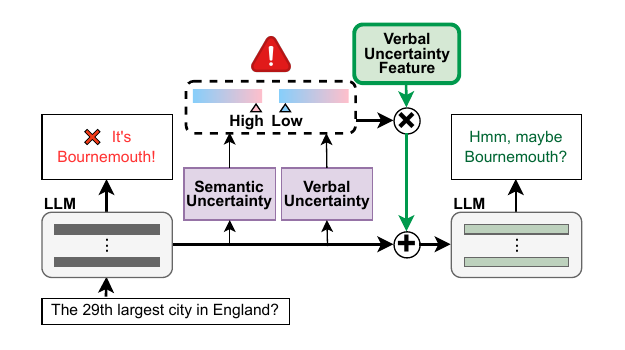}
        \caption{Illustration of our framework. We discover a linear verbal uncertainty feature (VUF) that controls model expression of uncertainty, and apply this insight to: 
    (1) detect hallucinations arising from the miscalibration between high semantic uncertainty (SU) and low verbal uncertainty (VU); 
    (2) mitigate hallucinations by intervening on activations along the VUF direction at inference time to make VU more aligned with model's SU.
     % from middle to the last layers 
    % Here, pink represents high uncertainty value, while blue represents low value.
    For example, when asked "What is the 29th largest city in England?", the model initially responds with "It's Bournemouth", exhibiting high SU and VU.
    % instead of the correct answer "Milton Keynes." 
By applying the VUF to intervention, we improve the VU to better align with the SU and the response becomes "Hmm, maybe Bournemouth?" -- demonstrating a nuanced expression of uncertainty.}
        \label{fig:muc-diagram}
    \end{minipage}
    \hfill
    \begin{minipage}[t]{0.35\linewidth}
        \centering
        \includegraphics[width=\linewidth]{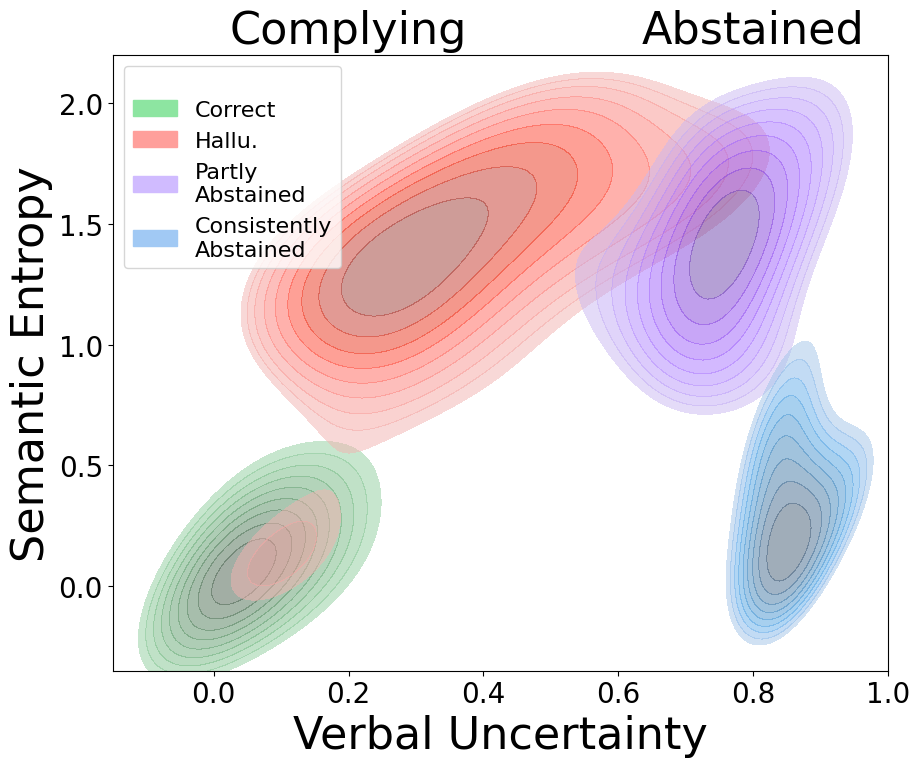}
        \caption{\textbf{Evidence of verbal-semantic uncertainty miscalibration.} This plot presents the Kernel Density Estimation (KDE) for samples from TriviaQA, categorized into four classes. These classes are based on the correctness of the answers generated by Llama3.1 and the consistency in abstaining.
    Miscalibration is indicated by high Semantic Entropy (proxy for SU) \& low VU in hallucinated answers (red), and low SU \& high VU in consistently abstained answers (blue).}
        \label{fig:se_in_4cases}
    \end{minipage}
\end{figure}

Our work begins with the analysis of the model representation space. We show that, similarly to refusal~\citep{arditi2024refusal} and other behaviours~\citep{zou2023representation}, the degree of uncertainty expressed by a model is mediated by a single direction, which we call the ``Verbal Uncertainty Feature'' (VUF). Specifically, we show that the hidden states of input questions answered with low and high  verbal uncertainty can be linearly separated.
 This allows us to use contrastive pairs \citep{burnsdiscovering,panickssery2023steering} to identify a single difference-in-means direction that can be intervened upon to control model expression of uncertainty. 

We next leverage our findings to study hallucinations through the lens of uncertainty features. 
As illustrated in  Figure \ref{fig:muc-diagram}, we highlight the misalignment between verbal uncertainty and the uncertainty about what meaning to convey in model outputs, i.e. semantic uncertainty, as an important factor contributing to hallucinations. We propose a novel method to detect model hallucinations that incorporates both verbal and semantic uncertainty information, and show that it outperforms detection methods that rely solely on semantic uncertainty. 
Next, we propose a method, \textbf{M}echanistic \textbf{U}ncertainty \textbf{C}alibration (MUC), that steers LLM activations using VUF to calibrate verbal uncertainty with the model semantic uncertainty. 

We demonstrate that MUC effectively reduces model confident hallucinations, achieving an average relative reduction of 29.6\% in short-form QA tasks. It also induces nuances expressions of uncertainty and achieves a 28.4\% improvement in the alignment between verbal and semantic uncertainty.
% 31.9\%  28.0\%

% Then we highlight the misalignment between linguistic uncertainty and the uncertainty of meaning of model outputs, i.e. semantic uncertainty, and propose novel method to detect hallucinations by incorporating both of these uncertainties. Our findings show the  improvement in the detection of hallucinations.

% Finally, we show that our findings of LUFs can be leveraged to steer the model towards a desired degree of caution. This enables calibrating a model's generation to align with its uncertainty regarding its knowledge via inference-time interventions, in a novel method that we call Mechanistic Uncertainty Calibration and builds over the existing literature on model steering \citep{hernandez2023inspecting, stickland2024steering, ravfogel2020null, hong2024intrinsic}. This approach reduces hallucinations in short-form generations by \nicola{Hallucination rate drop here} and induces a model to express nuanced uncertainty about its answers. Our method achieves an improvement in the correlation between verbal and intrinsic uncertainty, making models' answers more trustworthy. By expressing nuanced uncertainty, the model fosters user trust through transparency and honesty, implicitly directing users to cross-validate answers using other resources when appropriate.

% Our work demonstrates that insights gained from interpreting model internals can be practically beneficial, both for diagnosing model outputs and enhancing their quality for end users.
Our main contributions are therefore threefold: 
\begin{enumerate}[itemsep=-0.5ex]
\item We discovered that verbal uncertainty is mediated by a single direction in representation space, i.e. a \textbf{linear Verbal Uncertainty Feature} (VUF) (\S~\ref{sec:vuf}).
\item We \textbf{detect hallucinations} arising from the misalignment between high semantic and low verbal uncertainty by integrating both types of uncertainty (\S~\ref{subsec:detect}).
\item We introduce Mechanistic Uncertainty Calibration (MUC), an inference-time intervention mechanism using VUFs to \textbf{calibrate} verbal uncertainty with semantic uncertainty, thereby mitigating hallucinations (\S~\ref{subsec:mitigate}). 
\end{enumerate}

In addition, we introduce methods to quantify verbal uncertainty and metrics that help characterize the calibration between the two kinds of uncertainty without requiring the model to output numerical confidence estimates. Overall, this work contributes to a better understanding of LLMs, shows how to reduce hallucinations and thereby make LLMs more trustworthy.

\section{Background and Motivation}
\label{section:motivation}
The miscalibration of semantic and verbal uncertainty triggers overconfident hallucinations. To bring the discussion in a quantitative framework, we introduce in this section some definitions and measures.

% \subsection{Definition}
\subsection{Semantic Uncertainty}
\textit{Semantic Uncertainty} (SU) refers to the intrinsic uncertainty of an agent in the semantic meaning expressed by a statement. It reflects the confidence level of a model's prediction, focusing on its meaning and disregarding paraphrastic variations~\citep{lin2022teaching,kadavath2022language,mielke-etal-2022-reducing}. 
% focusing on semantic diversity and consistency rather than lexical diversity

We follow~\citet{kuhn2023semantic} in operationalizing this concept in terms of \textit{Semantic Entropy} that can be calculated as follows: for a given question, one first samples multiple model answers and clusters them into semantically equivalent groups, within which the meaning of every answer can be inferred from each other. Semantic entropy is then computed as the entropy of the probability distribution over semantic equivalence answer classes.

\subsection{Verbal Uncertainty}
\label{subsec:verbal_u}
% \textit{Verbal Uncertainty} (VU) refers to the degree of uncertainty in a statement that a speaker expresses through their choice of word and sentence structures.

\textit{Verbal Uncertainty} (VU) quantifies the degree of doubt a speaker expresses about a proposition \( P \), either explicitly or implicitly (e.g., "I doubt...", "Possibly..."). We formally define it as the complement of the subjective probability  a listener would associate with \( P \), conditioned on the utterance \( U \) and contextual information \( C \): 
% \vspace{-0.5em}
\begin{equation}
\text{VU}(U \mid C) = 1 - Pr(P \mid U, C)
\label{equ:vu}
\end{equation}
% the speaker

% Given a speaker S producing an utterance U conveying a proposition P to a listener L in context C, the \textit{Verbal Uncertainty (VU)} of U is inversely proportional to the probability according to L that S believes P, based only on U and C: 
% \vspace{-0.5em}
% \begin{equation}
% VU(U|C) = 1 - Pr_L(B(S,P)|U,C)
% \vspace{-0.5em}
% \label{equ:vu}
% \end{equation}
% where $B(S,P)$ denotes the fact that S believes P.} 
% We define verbal uncertainty as maximal when $U$ states that $S$ has insufficient knowledge to convey any relevant proposition $P$.
In the specific case of short-form QA, this definition can be instantiated with $U$ being the answer given by an agent in response to a question $C$.

%Answer A1 (e.g. "Hmm, maybe Bournemouth") is more verbally uncertain than A2 (e.g. "Bournemouth") if a listener would conclude that the agent believes the proposition P (e.g. "Bournemouth is the 29th largest city in England") more probable if given answer A2 than if given answer A1:
% $ Pr_L(B(S,$"Bournemouth is the 29th largest city in England")|"Bournemouth","What is the 29th largest city in England?"$) > $
% $Pr_L(B(S,$"Bournemouth is the 29th largest city in England")|"Hmm, maybe Bournemouth","What is the 29th largest city in England?") }

% [ncan] 
Answer $U_1$ is more verbally uncertain than $U_2$ if a listener would conclude that proposition \(P\) is more probable based on answer $U_2$ than based on $U_1$:
\[Pr(P|U_2,C) > Pr(P|U_1,C)\]
\noindent
where, e.g.:
\begin{itemize}
  \setlength{\itemsep}{0pt} % Adjusts space between items
  \setlength{\parskip}{0pt} % Adjusts space between paragraphs
  \setlength{\parsep}{0pt}  % Adjusts space between paragraphs within an item
    \item $P$: "Bournemouth is the 29th largest city in England"
    \item $C$: "What is the 29th largest city in England?"
    \item $U_1$: "Hmm, maybe Bournemouth"
    \item $U_2$: "Bournemouth"
\end{itemize}

% Linguistic research suggests that verbalized uncertainty recognition is highly subjective in nature \citep{giannakidou1999affective,de2012did}, 
We follow recent work in expression decisiveness quantification and employ "LLM-as-a-Judge" to measure VU \citep{yona-etal-2024-large, zheng2023judging}. 
% As shown in Figure~\ref{fig:vu_calculation}, 
Specifically, we sample multiple answers for each question and prompt an auxiliary evaluator LLM to directly assign a VU score to each answer. The VU for a question is the average of VU scores of all answers.~\footnote{See Appendix~\ref{app:prompts} for the prompt used for VU estimation.}
% which conveys how certain the language model is about the truthfulness of its answer
This approach has been shown to produce reliable uncertainty estimates that are highly correlated with human judgments of perceived assertiveness~\citep{yona-etal-2024-large,fagen2023perception}. To further validate the robustness of "LLM-as-a-Judge", we compute sentence embedding cosine similarities with predefined prototypical uncertainty expressions and find a high correlation with VU scores returned by LLMs (see Appendix~\ref{app:esu_euu} for details).

\begin{figure*}[]
    \centering
    \includegraphics[width=\textwidth]{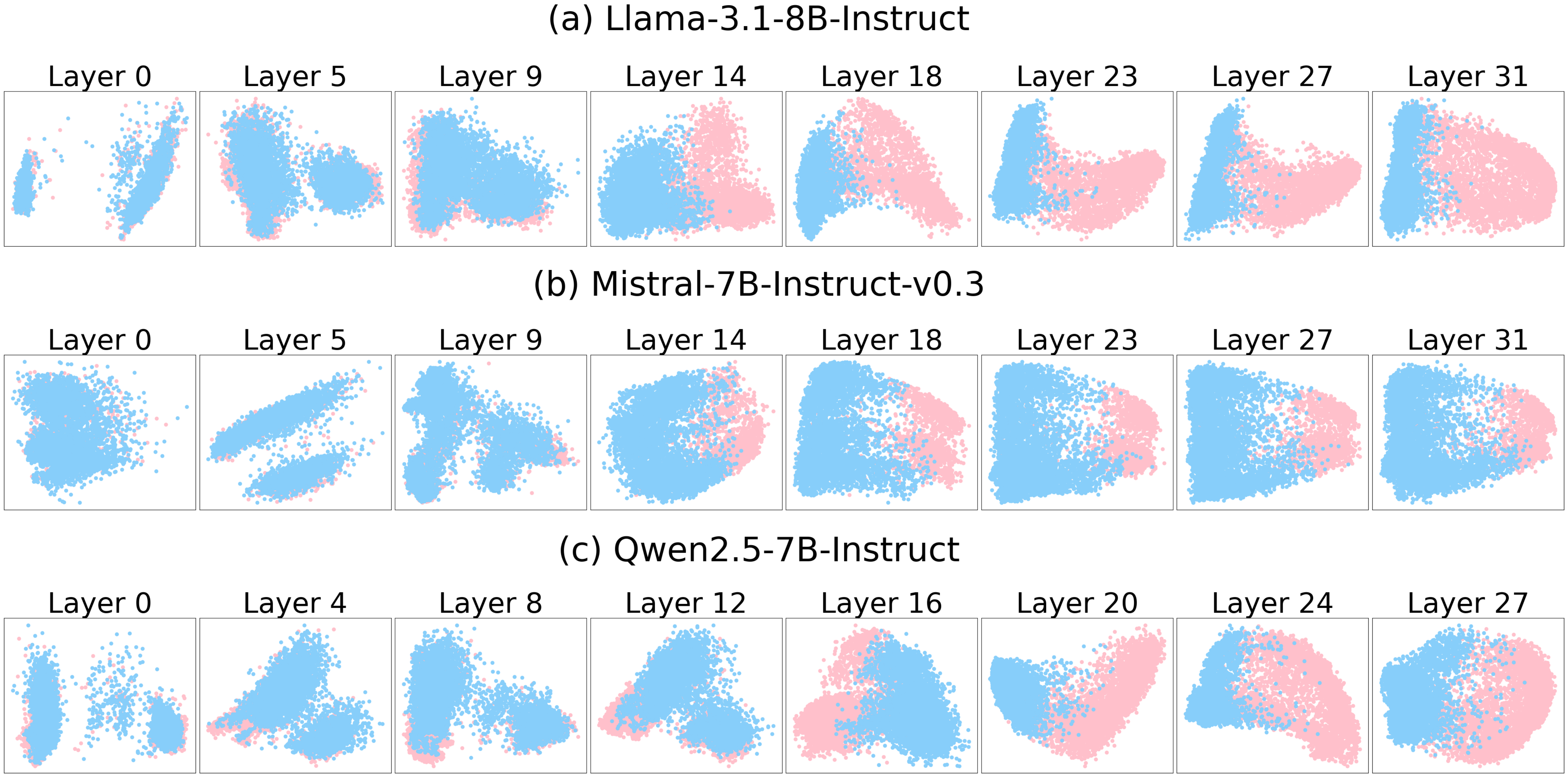}
    \caption{Visualization of verbalized certain (blue) vs. uncertain (pink) query representations exacted from selected layers of (a) Llama-3.1-8B-Instruct, (b) Mistral-7B-Instruct-v0.3, and (c) Qwen2.5-7B-Instruct on three datasets (TriviaQA, NQ-Open, PopQA). 
    Please refer to Appendix~\ref{app:vuf} for the visualization of representations exacted from all layers.}
    \label{fig:selected_linear_lu_Llama}
\end{figure*}

% \subsubsection*{Semantic Uncertainty}
% the advantage of Semantic Uncertainty-based hallucination detection}
\subsection{Hallucination arise from Miscalibration between Semantic and Verbal Uncertainty}
% \textbf{current problem of this section: we mix this section as both motivation for detection (using SU+VU) and also problem of miscalibrated hallucination.}
Ideally, VU should align with SU to let the model faithfully express uncertainty in the semantic meaning of its outputs.
However, observations indicate that the two types of uncertainty are not always correlated, resulting in hallucinations.
In this section, we quantitatively investigate and demonstrate the miscalibration between semantic and verbal uncertainty by analyzing samples from TriviaQA using Llama3.1~\footnote{See Appendix~\ref{app:examples_4cases} for examples.}.
% We analyze samples from TriviaQA (Figure \ref{fig:se_in_4cases}) to investigate the correlation between these uncertainties.

Following~\citet{kossen2024semantic,farquhar2024detecting}, for each question, we generate a response using a low temperature (0.1) to obtain the most likely answer, and then sample multiple responses using a high temperature (1.0).
We categorize the samples into two primary groups based on the VU level of the most likely answer: 
those that include abstentions (abstained) and those that do not (complying). We further subdivide these categories. 
For complying responses, we assess whether the answers are hallucinated or correct (hallucinated/correct).
For abstained, we determine if the model consistently refuses (i.e., all samples), or if it complies at least once among the multiple sampled answers (``partly abstained'')~\footnote{We obtain the MANOVA results which indicate significant differences in 4 groups with p-values < 0.0001.}. 
% with Wilks' Lambda = 0.4973, Pillai's Trace = 0.5672, Hotelling-Lawley Trace = 0.8813, and Roy's Greatest Root = 0.6947, all 

 % To strongly justify increasing VU but stopping short of abstention we should show that there is a progression, with more semantic uncertainty correlating with a higher hallucination rate, and a worse correlation between VU before intervention and hallucination rate. In this way we could claim that our approach optimizes the trade-off balance between hallucinations and false refusals.
 
% high VU high SU partly abstain
% low VU low SU correct
% low VU high SU hallucinated
% high VU low SU consitently refuse
% In complying responses characterized by low VU, consistently providing the correct answer yields low SU, whereas hallucinations result in high SU.
% In the abstained responses with high VU, consistent abstention yields low SU, while partial abstention leads to high SU.
% These observations highlight a \textbf{miscalibration} between VU and SU in two scenarios: consistent abstention (high VU \& low SU) and "confident hallucinated" responses (low VU \& high SU). The latter scenario is the focus of this work, where the model neither abstains nor expresses uncertainty while providing inaccurate and hallucinated answers.

As shown in Figure \ref{fig:se_in_4cases}, abstained responses have high VU, which is expected. Consistently abstained ones have low SU, but this is not a problematic mismatch, rather an artifact of using semantic entropy as a proxy for SU: these are cases where the model ``knows that it does not know'' and deals with them appropriately.
There is however a large segment of complying answers with high SU and low VU that are hallucinations~\footnote{There are outlier hallucinations with low SU. For a detailed analysis, please refer to Appendix~\ref{app:low_SU_hallucinations}.}: this is the focus of our intervention. 
We show in \S~\ref{subsec:detect} that combining predictions of VU and SU helps detecting hallucination. 
Moreover, we show in \S~\ref{subsec:mitigate} that modulating VU to better reflect SU is crucial to prevent confident hallucinations and optimize the trade-off balance between hallucinations and false abstention.

% Semantic uncertainty standalone cannot distinguish between consistent abstention and correct answers, as well as between partial abstention and non-abstained hallucinated responses. Therefore, combining both semantic and verbal uncertainties is necessary for effective hallucination detection, as introduced in \S~\ref{subsec:detect}. Here, abstained answer is not considered to be hallucination although it cannot provide correct answer. 

% Semantic uncertainty struggles to effectively distinguish between consistent abstention and correct answers, as well as between partial abstention and non-abstained hallucinated responses. Therefore, combining both uncertainties is necessary for effective hallucination detection, as introduced in \S~\ref{subsec:detect}.
% Based on the insights, considering both uncertainties is necessary for effective hallucination detection introduced in \S~\ref{subsec:detect}.
% Moreover, calibrating verbal with SU is crucial to prevent "confident hallucination" introduced in \S~\ref{subsec:mitigate}.

\subsection{Semantic Space of LLM}
\label{subsec:semantic_space}
% \bang{Should we add the motivation of exploring VUF here? (the first paragraph in section 3)}
Recent research supports the hypothesis that language models represent features or concepts as linear directions within their activation space~\citep{mikolov2013linguistic, bolukbasi2016man, elhage2022toy, park2023linear,ferrando2024know}. These features include attributes such as harmlessness \citep{wolf2024tradeoffs, arditi2024refusal}, truthfulness \citep{marks2023geometry, li2024inference}, sentiment \citep{tigges2023linear}, and language \citep{bricken2023monosemanticity}. 
Building on these insights, we investigate the linear representation of VU, aiming to validate its representation and control its level.

\section{Verbal Uncertainty Feature (VUF)}
\label{sec:vuf}

In this section we show that verbal uncertainty is mediated by a single direction.

\subsection{Feature Extraction}
\label{subsec:luf_feature_extract}

% \noindent \textbf{Linguistic Uncertainty Direction}
To identify the verbal uncertainty features (VUFs) in the model's residual stream activations, we adopt the difference-in-means technique \citep{Belrose_2023}, which has been shown to effectively disentangle key feature information \citep{panickssery2023steering, marks2023geometry, arditi2024refusal, yu2024robust}.

We first collect a set of pairs of questions and generated answers for which the model generates answers with high VU, denoted as question $x \in \mathcal{D}_{uncertain}$, and another set for questions on which the model generates certain answers with low VU, denoted as $x \in \mathcal{D}_{certain}$. We construct $\mathcal{D}_{uncertain}$ and $\mathcal{D}_{certain}$ by choosing top $N_{uncertain}$ and bottom $N_{certain}$ pairs sorted by their VU score, which is calculated using the LLM-as-a-Judge method described in \S~\ref{subsec:verbal_u} with the Llama3.1-70B-Instruct model \citep{dubey2024llama}.
To generate answers, we use the ``uncertainty'' eliciting prompt provided in Appendix \ref{appendix:prompt_lu}. 
We then calculate the difference between the model’s mean last-token residual stream activations $\mathbf{h}^{(l)}(x)$ for each layer $l$ when processing these two sets of questions. This difference is then L2 normalized, as illustrated in the following equation:
\vspace{-0.5em}
\begin{equation}
\label{equation: luf}
\begin{split}
\hat{\mathbf{r}}_{\text{VU}}^{(l)} = 
& \, \frac{1}{|\mathcal{D}_{\text{uncertain}}|} \sum_{x \in \mathcal{D}_{\text{uncertain}}} \mathbf{h}^{(l)}(x)  \\
& - \frac{1}{|\mathcal{D}_{\text{certain}}|} \sum_{x \in \mathcal{D}_{\text{certain}}} \mathbf{h}^{(l)}(x)
\end{split}
\end{equation}

\vspace{-0.5em}
\begin{equation}
\label{equation: luf2}
\mathbf{r}_{\text{VU}}^{(l)} = \frac{\hat{\mathbf{r}}_{\text{VU}}^{(l)}}{\|\hat{\mathbf{r}}_{\text{VU}}^{(l)}\|}
\end{equation}
\vspace{-0.5em}

\begin{figure}[]
    \centering
    \includegraphics[width=0.6\textwidth]{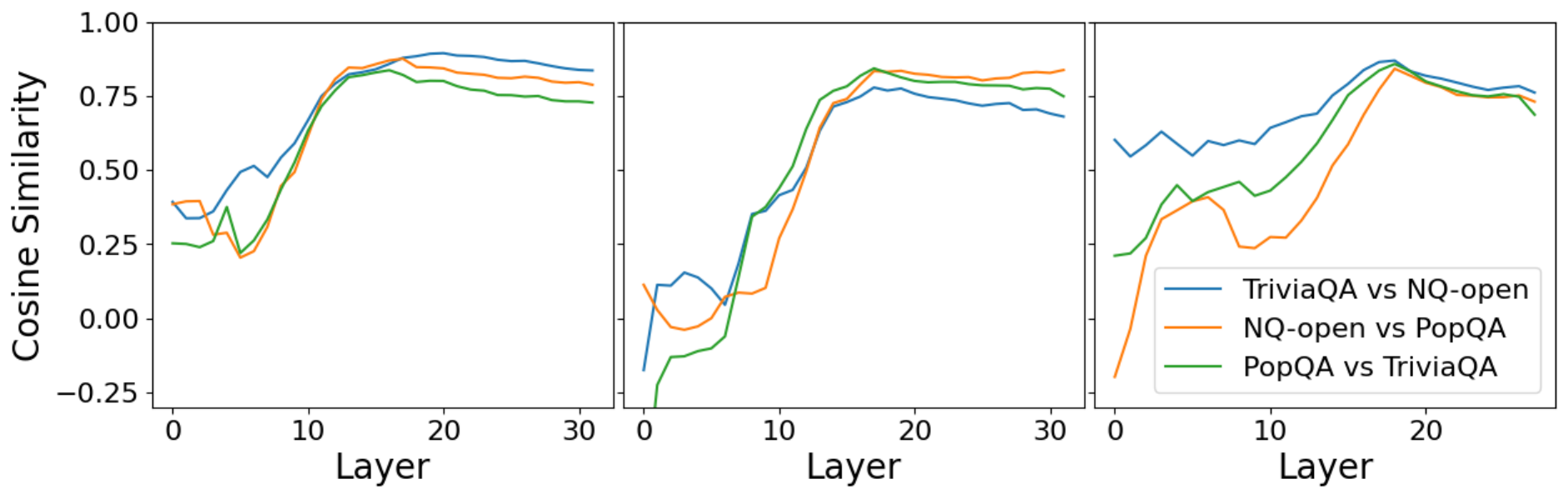}
    \caption{Compare VUFs exacted from different datasets from Llama-3.1-8B-Instruct, Mistral-7B-Instruct-v0.3, and Qwen2.5-7B-Instruct}
    \label{fig:cos_sim}
\end{figure}

\begin{figure}[ht]
    \centering
    \includegraphics[width=0.6\textwidth]{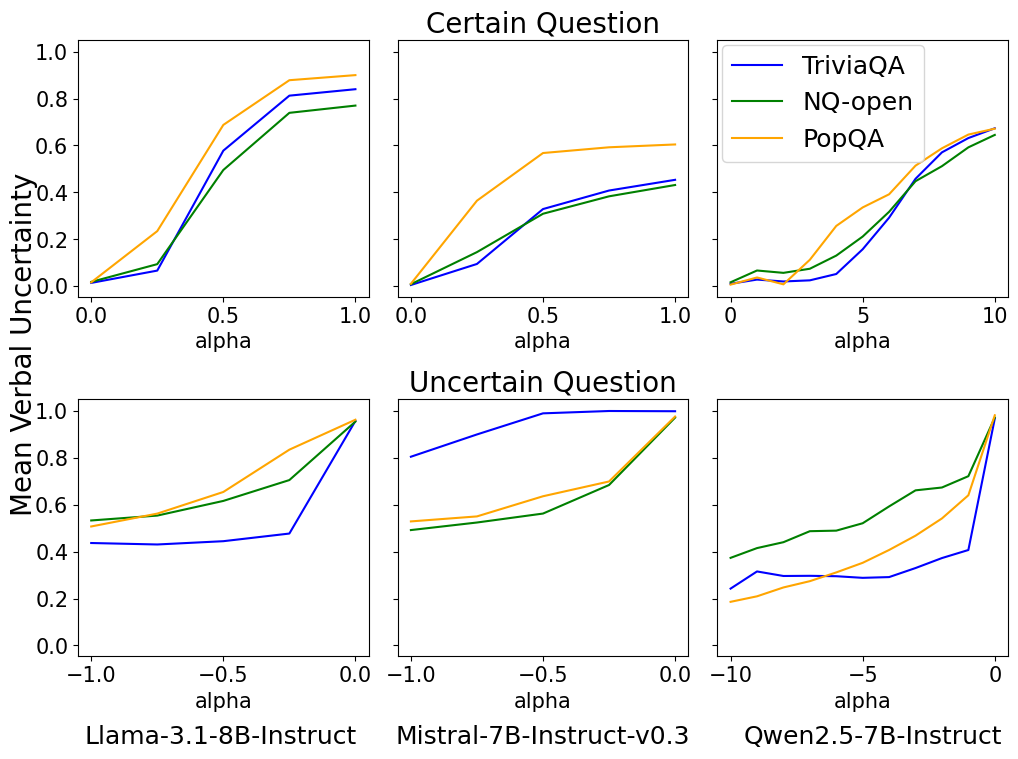}
    \caption{Mean model-generated answer verbal uncertainty on three QA datasets with varying degrees of inference-time VUF intervention (modulated by the intervention intensity $\alpha$). }
    \label{fig:causal_lu}
\end{figure}

\subsection{Discovery of Linear Verbal Uncertainty Features}
\label{subsec: discovery of vuf}
To empirically demonstrate the VUFs explained above, we adopt three closed-book short-form QA datasets: TriviaQA~\citep{joshi2017triviaqa}, NQ-Open ~\citep{kwiatkowski2019natural}, and PopQA~\citep{mallen2023llm_memorization};
% For generating responses and quantifying uncertainties based on their responses and hidden states, we employ 
and consider the following models: Llama3.1-8B-Instruct~\citep{dubey2024llama}, Mistral-7B-Instruct-v0.3~\citep{jiang2023mistral}, and Qwen2.5-7B-Instruct~\citep{yang2024qwen2}
~\footnote{For simplicity, we will refer to them as Llama3.1, Mistral, and Qwen2.5, respectively.}.

\paragraph{Visualization}
We extract the activations of the last token for each question at each layer from $\mathcal{D}_{uncertain}$ and $\mathcal{D}_{certain}$ and project them into a 2D space using PCA.
As illustrated in Figure \ref{fig:selected_linear_lu_Llama}, the activations of samples labeled as "certain" and "uncertain" are clearly linearly separated into two distinct clusters, \textit{starting from the middle layers}. 
This strongly indicates that $\mathbf{r}_{\text{VU}}^{(l)}$ represents a meaningful linear direction that reflects the VU level of questions in hidden states. We refer to $\mathbf{r}_{\text{VU}}^{(l)}$ as VUFs.

% We compute the first two principal components of activations at each layer of two sets and project onto 2D space which can be seen in Figure \ref{fig:pca_trivia_llm_llama}. We can clearly see that activations of samples selected as "certain" and "uncertain" are linearly separated into two clusters starting from the middle layers. It strongly suggests  \textit{the existence of linguistic uncertainty features (LUFs)}. This observation is shown with other datasets and linguistic uncertainty score functions as seen in Appendix \ref{app:pca}.

\paragraph{Effective Layer Selection}
% Cross-Dataset 
\label{paragraph: layer consistency}
To identify the effective layers of VUFs, we analyze VUFs obtained from each layer of three different models.
We measure the cosine similarity of distinct VUFs extracted from TriviaQA, NQ-Open, and PopQA datasets respectively.
The results presented in Figure \ref{fig:cos_sim} show a high cosine similarity between VUFs from different datasets, particularly in \textit{the middle and subsequent layers}. 
This pattern is aligned with visualization and consistent across all models and datasets we examined.
Observations from both visualization and similarity across datasets indicate that reliable VUFs are best extracted from the middle to the last layers.

\paragraph{Causal Validation}
\label{subsec: luf_causal_validation}
We validate the causal connection between VUFs and the model's VU by analyzing the generation behavior as we modulate the strength of the corresponding feature through simple linear interventions.
% we investigate control degrees of linguistic uncertainty in model responses.
Inspired by \citet{li2024inference}, we intervene on model activations of all tokens by steering them along a set of VUF directions.

For each layer $l$, we extract VUs $\mathbf{r}_{\text{VU}}^{(l)} \in \mathbb{R}^{d_{model}}$.
Specifically, the VU feature vector $\mathbf{r}_{\text{VU}}^{(l)}$ serves as a directional guide for steering activations, as described in the equation below:
\vspace{-0.5em}
\begin{equation}
h^{(l)}(x) \leftarrow h^{(l)}(x) + \alpha * \mathbf{r}_{\text{VU}}^{(l)}
\label{equ:iti}
\end{equation}
where $\alpha$ is the intensity of intervention, and $\mathbf{r}_{\text{VU}}^{(l)}$ is the verbal uncertainty feature at layer $l$.  
The results presented in Figure \ref{fig:causal_lu} show that adding VUFs to model activations ($\alpha>0$) increases the VU of the model outputs. Conversely, removing VUFs from activations ($\alpha<0$) decreases this uncertainty.
As the intensity of VUFs ($|\alpha|$) gets stronger, the VU scores exhibit greater changes. This trend remains consistent across all models and datasets we studied. This show the potential of VU calibration in model generation. 
We will further explore how to leverage this phenomenon in \S \ref{subsec:mitigate}.

Interestingly, although Qwen2.5 exhibits a similar trend, it is significantly less sensitive compared to Llama3.1 and Mistral. 
This is due to the normalization of the VUF. Qwen embeddings have larger norms, resulting in longer distances between clusters.

To address potential circularity concerns due to the double use of LLM in VUF extraction and VU evaluation, we validate our findings using an alternative VUF extraction method, detailed in Appendix~\ref{app:circularity}.

\begin{figure}[]
    \centering
    \includegraphics[width=0.6\textwidth]{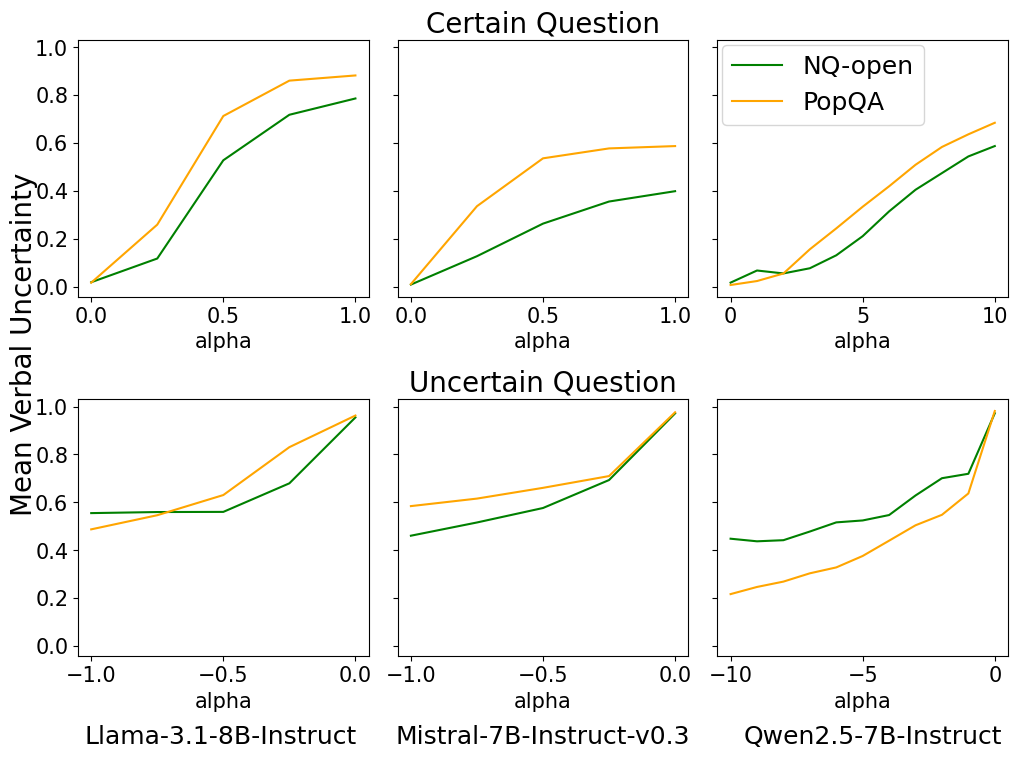}
    \caption{Causal Validation on NQ-Open and PopQA with the VUF exacted from the OOD dataset TriviaQA. }
    \label{fig:ood_causal_lu}
\end{figure}

\paragraph{VUFs are Consistent Across Different Datasets}
\label{paragraph: luf consistent across datasets}
To investigate the generalization of VUF across datasets, we use VUFs extracted from TriviaQA to control the VU level of other datasets: NQ-Open and PopQA.
As shown in Figure~\ref{fig:ood_causal_lu}, adding or removing TriviaQA VUFs increases or decreases the VU of model outputs for these datasets.
These two findings indicate that VUFs are consistent across different datasets, suggesting that a universal VUF can be derived and utilized in our experiments further in the paper. 
Similar results using other VU scores are provided in Appendix~\ref{app:lu_data_sim}.

Therefore, once we identify the appropriate layers for each model, these selections remain consistent across different datasets, eliminating the need to repeat the selection process for other datasets.

\section{Verbal Uncertainty and Hallucination}
\label{sec:application}
Confident hallucinations arise from a miscalibration between VU and SU, where the model fails to express its high uncertainty in its generated output. Taking advantage of this miscalibration, we can detect hallucinations (\S~\ref{subsec:detect}).
Furthermore, we mitigate confident hallucinations by calibrating these two uncertainties using VUFs discovered in \S~\ref{sec:vuf} (\S~\ref{subsec:mitigate}).
% Based on the analysis of uncertainty miscalibration \S~\ref{section:motivation} and the discovery of VUF in \S~\ref{sec:vhf}, this section focuses on applying uncertainties to detect and mitigate hallucinations.

% In this work, we focus on detecting "confident hallucinations," which are samples characterized by low VU and high SU. These are considered the most detrimental type of error among the four cases previously discussed. Our detection task aims to identify these hallucination instances within sentence-format, closed-book QA scenarios.
% We define samples with low VU and high SU as "confident hallucinations," which are considered the most detrimental type of error among the four cases previously discussed.
% The objective of our detection task is to identify these hallucination instances within sentence-format, closed-book QA scenarios.

\subsection{Hallucination Detection with Semantic and Verbal Uncertainty}
\label{subsec:detect}
\begin{table}[]
\centering
\begin{minipage}{0.48\linewidth}
\centering
\resizebox{\linewidth}{!}{
\begin{NiceTabular}{rrcc|cc|cc}
\toprule
\multicolumn{1}{c}{} & \multicolumn{1}{l}{} & \multicolumn{2}{c}{\textbf{Llama}} & \multicolumn{2}{c}{\textbf{Mistral}} & \multicolumn{2}{c}{\textbf{Qwen}} \\
\multicolumn{1}{c}{\multirow{-2}{*}{\textbf{Dataset}}} & \multicolumn{1}{c}{\multirow{-2}{*}{\textbf{Feature}}} & \textbf{AUROC} & \textbf{ACC} & \textbf{AUROC} & \textbf{ACC} & \textbf{AUROC} & \textbf{ACC} \\ \midrule
& Semantic & 79.21 & \textbf{81.1} & 71.51 & 74.6 & 72.18 & 72.0 \\
 & Verbal & 72.1 & 80.1 & 68.20 & 72.4 & 72.8 & 72.7 \\
\multirow{-3}{*}{\textbf{TriviaQA}} & Combined & \textbf{79.71} & 80.8 & \textbf{72.99} & \textbf{74.6} & \textbf{74.71} & \textbf{72.7} \\ \midrule
 & Semantic & 65.29 & 70.7 & 64.47 & 62.1 & 56.74 & 53.6 \\
 & Verbal & 54.04 & 71.2 & 65.26 & 62.5 & 61.85 & \textbf{60.7} \\
\multirow{-3}{*}{\textbf{NQ-Open}} & Combined & \textbf{66.02} & 70.3 & \textbf{68.96} & \textbf{64.8} & \textbf{62.36} & 58.7 \\ \midrule
 & Semantic & 71.16 & \textbf{81.2} & 66.03 & 71.1 & 53.44 & 71.7 \\
 & Verbal & 62.30 & 81.0 & 71.13 & 71.7 & \textbf{75.66} & \textbf{76.2} \\
\multirow{-3}{*}{\textbf{PopQA}} & Combined & \textbf{75.66} & 81.1 & \textbf{73.82} & \textbf{75.7} & 75.43 & 76.0 \\
\bottomrule
\end{NiceTabular}
}
\caption{Detection Results based on Uncertainty for Llama-3.1-8B-Instruct, Mistral-7B-Instruct-v0.3, and Qwen2.5-7B-Instruct}
\label{tab:detect_results_3models}
\end{minipage}
\hfill
\begin{minipage}{0.48\linewidth}
\centering
\resizebox{\linewidth}{!}{
\begin{tabular}{ccc|cc|cccc}
\hline
\multirow{3}{*}{\textbf{Dataset}} & \multicolumn{2}{c|}{\multirow{2}{*}{\textbf{SEP}}} & \multicolumn{2}{c|}{\multirow{2}{*}{\textbf{EigenScore}}} & \multicolumn{4}{c}{\textbf{Our method}} \\
 & \multicolumn{2}{c|}{} & \multicolumn{2}{c|}{} & \multicolumn{2}{c}{\textbf{Calculated}} & \multicolumn{2}{c}{\textbf{Probe-Predicted}} \\
 & \textbf{AUROC} & \textbf{ACC} & \textbf{AUROC} & \textbf{ACC} & \textbf{AUROC} & \textbf{ACC} & \textbf{AUROC} & \textbf{ACC} \\ \hline
\textbf{TriviaQA} & 66.85 & 66.0 & 64.83 & 53.5 & \textbf{79.71} & \textbf{80.8} & 73.53 & 80.1 \\
\textbf{NQ-Open} & 54.07 & 53.9 & 56.29 & 49.3 & \textbf{66.02} & 70.3 & 57.15 & \textbf{71.3} \\
\textbf{PopQA} & 70.17 & 65.6 & 59.33 & 44.9 & \textbf{75.66} & \textbf{81.1} & 74.76 & 81.0 \\ \midrule
\end{tabular}
}
\caption{Detection Results on Llama-3.1-8B-Instruct. `Calculated' means that the SE feature is computed after sampling multiple answers, 'Probe-Predicted' means that SE is as predicted by a probe that takes as input the embeddings of the last token of the question, and therefore does not require sampling.}
\label{tab:ablation_detect_Llama}
\end{minipage}
\end{table}

\label{subsec:our_detect_method}
% \item\textbf{Detection with Regressor Uncertainty Probe:} 
We propose to detect hallucinations by leveraging both verbal and semantic uncertainties. Our approach utilizes a simple logistic regression model to predict the presence of hallucinations. We demonstrate that combining VU with SU significantly enhances the detection performance.

Measuring Semantic Entropy, our proxy for SU, requires generating multiple samples and running auxiliary models~\citep{farquhar2024detecting}.
We therefore also consider training Uncertainty Probes for uncertainty quantification following~\citep{kossen2024semantic} to ensure cost-efficiency.
These probes are linear models trained on the hidden states of LLMs to predict numerical uncertainty values. The hidden states are extracted from the last token of the question and sourced from multiple layers within the LLM~\footnote{Please refer to Appendix~\ref{app:experiment_detail_detect} for implement details.}.
During testing, the input to the logistic regression model consists of predicted verbal and semantic uncertainties obtained from two regressor probes.\footnote{Probes trained as binary classifiers over thresholded continuous values perform similarly, see Appendix~\ref{app:classifier_probe}.} 
% Instead of using calculated scores, we utilize the scores predicted by the Regressor Uncertainty Probe.

\begin{table*}[]
\centering
\resizebox{\linewidth}{!}{
\begin{tabular}{l|ccc|cc|cc|cc|cc|cc|cc}
\toprule
\multicolumn{1}{c|}{\multirow{3}{*}{\textbf{Dataset}}} & \multicolumn{3}{c|}{\textbf{Hallucination Rate↓}} & \multicolumn{2}{c|}{\multirow{2}{*}{\textbf{Correct. Rate↑}}} & \multicolumn{2}{c|}{\multirow{2}{*}{\textbf{Refusal Rate}}} & \multicolumn{2}{c|}{\multirow{2}{*}{\textbf{\begin{tabular}[c]{@{}c@{}}VU/SU\\ Disagree. Rate ↓\end{tabular}}}} & \multicolumn{2}{c|}{\multirow{2}{*}{\textbf{Correlation↑}}} & \multicolumn{2}{c|}{\multirow{2}{*}{\textbf{\begin{tabular}[c]{@{}c@{}}VU for\\ Incorrect ↑\end{tabular}}}} & \multicolumn{2}{c}{\multirow{2}{*}{\textbf{\begin{tabular}[c]{@{}c@{}}VU for\\ Correct\end{tabular}}}} \\
\multicolumn{1}{c|}{} & \textbf{Conf./Overall} & \textbf{Conf.} & \multicolumn{1}{c|}{\textbf{Overall}} & \multicolumn{2}{c|}{} & \multicolumn{2}{c|}{} & \multicolumn{2}{c|}{} & \multicolumn{2}{c|}{} & \multicolumn{2}{c|}{} & \multicolumn{2}{c}{} \\
 & before & after & after & before & after & before & after & before & after & before & after & before & after & before & after \\ \hline
\multicolumn{16}{c}{Llama3.1-8B} \\ \hline
TriviaQA & 23.3 & \textbf{19.0} & 21.2 & \textbf{71.3} & 70.6 & 5.4 & 8.2 & 21.50 & \textbf{21.40} & 0.59 & \textbf{0.63} & 0.50 & \textbf{0.55} & 0.16 & \textbf{0.16} \\
NQ-Open & 40.2 & \textbf{26.2} & 32.7 & \textbf{50.7} & 47.7 & 9.1 & 19.6 & 35.10 & \textbf{18.90} & 0.38 & \textbf{0.69} & 0.37 & \textbf{0.54} & 0.17 & \textbf{0.24} \\
PopQA & 33.7 & \textbf{21.6} & 23.2 & \textbf{23.5} & 21.0 & 42.8 & 55.8 & 50.70 & \textbf{44.70} & 0.05 & \textbf{0.34} & 0.61 & \textbf{0.73} & 0.17 & \textbf{0.20} \\
\rowcolor[HTML]{D9D9D9} 
Average & 32.4 & \textbf{22.3} & 25.7 & \textbf{48.5} & 46.4 & 19.1 & 27.9 & 35.80 & \textbf{28.30} & 0.34 & \textbf{0.55} & 0.49 & \textbf{0.61} & 0.17 & \textbf{0.20} \\ \hline
\multicolumn{16}{c}{Mistral-7B} \\ \hline
TriviaQA & 30.2 & \textbf{19.7} & 26.8 & \textbf{67.9} & 67.0 & 1.9 & 6.2 & 27.50 & \textbf{16.80} & 0.46 & \textbf{0.66} & 0.19 & \textbf{0.39} & 0.04 & \textbf{0.05} \\
NQ-Open & 52.2 & \textbf{40.8} & 46.9 & \textbf{41.7} & 39.4 & 6.1 & 13.7 & 46.80 & \textbf{19.80} & 0.24 & \textbf{0.58} & 0.23 & \textbf{0.40} & 0.07 & \textbf{0.10} \\
PopQA & 58.2 & \textbf{26.7} & 32.5 & \textbf{26.4} & 23.9 & 15.4 & 43.6 & 50.80 & \textbf{28.50} & 0.15 & \textbf{0.53} & 0.30 & \textbf{0.64} & 0.07 & \textbf{0.15} \\
\rowcolor[HTML]{D9D9D9} 
Average & 46.9 & \textbf{29.1} & 35.4 & \textbf{45.3} & 43.4 & 7.8 & 21.2 & 41.70 & \textbf{21.70} & 0.28 & \textbf{0.60} & 0.20 & \textbf{0.50} & 0.06 & \textbf{0.10} \\ \hline
\multicolumn{16}{c}{Qwen2.5-7B} \\ \hline
TriviaQA & 37.9 & \textbf{23.4} & 34.4 & \textbf{58.6} & 58.1 & 3.5 & 7.5 & 27.10 & \textbf{22.20} & 0.57 & \textbf{0.59} & 0.43 & \textbf{0.51} & 0.14 & \textbf{0.14} \\
NQ-Open & 61.6 & \textbf{46.8} & 56.5 & \textbf{30.4} & 30.1 & 8.0 & 13.4 & 44.50 & \textbf{32.50} & 0.31 & \textbf{0.38} & 0.39 & \textbf{0.46} & 0.18 & \textbf{0.19} \\
PopQA & 44.8 & \textbf{33.6} & 38.3 & \textbf{18.1} & 16.4 & 37.1 & 45.3 & 46.80 & \textbf{43.00} & 0.08 & \textbf{0.22} & 0.69 & \textbf{0.75} & 0.21 & \textbf{0.21} \\
\rowcolor[HTML]{D9D9D9} 
Average & 48.1 & \textbf{34.6} & 43.1 & \textbf{35.7} & 34.9 & 16.2 & 22.1 & 39.50 & \textbf{32.60} & 0.32 & \textbf{0.39} & 0.51 & \textbf{0.57} & 0.17 & \textbf{0.18} \\ \hline
\multicolumn{16}{c}{Llama3.1-70B} \\ \hline
TriviaQA & 12.1 & \textbf{10.1} & 11.8 & \textbf{87.0} & 86.8 & 0.9 & 1.4 & 7.5 & \textbf{7.1} & 0.71 & \textbf{0.80} & 0.29 & \textbf{0.35} & 0.06 & \textbf{0.07} \\
NQ-Open & 35.7 & \textbf{32.3} & 34.0 & \textbf{60.8} & 59.5 & 3.5 & 6.5 & 21.1 & \textbf{15.1} & 0.49 & \textbf{0.73} & 0.27 & \textbf{0.36} & 0.08 & \textbf{0.09} \\
PopQA & 41.4 & \textbf{28.0} & 35.2 & \textbf{44.6} & 42.4 & 14.0 & 22.4 & 22.2 & \textbf{14.8} & 0.59 & \textbf{0.75} & 0.48 & \textbf{0.62} & 0.17 & \textbf{0.18} \\
\rowcolor[HTML]{D9D9D9} 
Average & 29.7 & \textbf{23.5} & 27.0 & \textbf{64.1} & 62.9 & 6.1 & 10.1 & 16.9 & \textbf{12.3} & 0.60 & \textbf{0.76} & 0.35 & \textbf{0.44} & 0.10 & \textbf{0.11} \\
\bottomrule
\end{tabular}
}
\caption{Mitigation Results for Llama-3.1-8B-Instruct, Mistral-7B-Instruct-v0.3, Qwen2.5-7B-Instruct, and Llama-3.1-70B-Instruct. 
`Before' represents the original generation and `after' represents the generation after Mechanistic Uncertainty Calibration. The original generation is always confident, so there is no difference between `Confident' and `Overall'.}
\label{tab:mitigation}
\end{table*}

% \paragraph{Evaluation Metrics}
To evaluate the performance of hallucination detection, we follow~\citet{kossen2024semantic, orgad2024llms} and adopt the area under the receiver operating characteristic curve (AUROC) as the main metric. Additionally, we use accuracy (ACC) as a reference metric.
% These metrics are computed by comparing the model outputs against gold labels indicating whether the responses include hallucinations or not. 

\paragraph{Baselines}
We adapt SEP~\citep{kossen2024semantic}, a hallucination detection method that trains a probe to predict binarized SU based on hidden states. We employ the sentence-form and TBG (token before generating) settings of SEP.
In contrast to the original setup, we classify abstained samples as non-hallucinated. 
% with high VU 
% Eigenscore~\citep{chen2024inside}

\paragraph{Result} 
As shown in Tab.~\ref{tab:detect_results_3models}, incorporating VU alongside SU improves detection performance for all models.
% Instead of calculated uncertainty scores, we use the uncertainty scores predicted by Regressor Uncertainty Probe during test. 
The accuracy when using probe-predicted uncertainties is similar to that obtained when using calculated SU (Tab.~\ref{tab:ablation_detect_Llama}). This is important because it means that it is possible to predict a high risk of hallucination already after the \textit{prefill} stage of decoding, before starting autoregressive generation.
% Additionally, we explore another detection method, the Classifier Binarized Uncertainty Probe, and show that VU also enhances hallucination detection in classifier settings (see Appendix~\ref{app:classifier_probe} for details).
% ~\footnote{Please refer to Appendix~\ref{app:classifier_probe} for details.}

\subsection{Hallucination Mitigation via Inference-time Mechanistic Uncertainty Calibration}
\label{subsec:mitigate}
\label{subsec:mitigate_method}
In \S~\ref{sec:vuf}, we observed the existence of universal VUFs extracted from the middle layers to the last. These VUFs enable us to modulate the degree of verbal uncertainty in model responses.
Building on the insights, we propose Mechanistic Uncertainty Calibration (MUC). This method leverages VUFs to calibrate verbal uncertainty with semantic uncertainty.

For each layer $l$, we extract VUF vectors from the last token of question, $\mathbf{r}_{\text{VU}}^{(l)} \in \mathbb{R}^{d_{model}}$.
% Specifically, the verbal uncertainty feature vector $\mathbf{r}_{\text{VU}}^{(l)}$ serves as a directional guide for steering activations, while the normalized semantic uncertainty determines the magnitude, as described in the equation below:
We then modulate the influence of these features through straightforward linear interventions on all tokens in detectd hallucinated responses. Specifically, $\mathbf{r}_{\text{VU}}^{(l)}$ serves as a directional guide for steering activations, as described in the equation:

\begin{equation}
h^{(l)}(x) \leftarrow h^{(l)}(x) + \alpha_\text{su}(x) * \mathbf{r}_{\text{VU}}^{(l)}
\label{equ:iti2}
\end{equation}
where the magnitude of intervention  
\begin{equation}
% \small
\alpha_\text{su}(x) = clip(su(x)_{norm} - vu(x), 0, max_\alpha)
\end{equation}
is the gap between min-max normalized SU and VU of input questions x, and $\mathbf{r}_{\text{VU}}^{(l)}$ is the VUF at layer $l$.  
Please refer to Appendix~\ref{app: expert detail mitigate} for details.

Unlike previous work on abstention~\citep{feng-etal-2024-dont, zhang-etal-2024-r}, i.e. which advocate declining to answer in the face of uncertainty, we aim to incorporate nuanced uncertainty in the output text form.
Our method does not involve fine-tuning of the model, additional system prompt design, or sampling methods required by previous mitigation works on nuanced uncertainty.
We show the existence of universal VUFs that can be pre-comuputed, which saves on computation overhead. 
Our method leverages the model's underutilized inherent ability to express nuanced uncertainty, allowing it to better manage and communicate its confidence levels.

\paragraph{Evaluation Metrics}
To evaluate the level of hallucination and the calibration of verbal and semantic uncertainties in generated responses, we use the following metrics:
\begin{itemize}
  \setlength{\itemsep}{0pt} % Adjusts space between items
  \setlength{\parskip}{0pt} % Adjusts space between paragraphs
  \setlength{\parsep}{0pt}  % Adjusts space between paragraphs within an item
\item Overall Hallucination Rate: The proportion of samples where the model doesn't refuse and the answer is not entailed by the golden answer.

\item Confident Hallucination Rate: The proportion of responses that are not entailed by the golden answer and have a low VU below a predefined threshold. The threshold is identified by minimizing the sum of squared distances from VU to the threshold~\citep{kossen2024semantic}.~\footnote{In the original output, confident hallucinations (incorrect and certain) and hallucinations (incorrect and non-refusal) are the same, as verbally uncertain answers without refusal are rare.}
\item Correctness Rate: The proportion of samples entailed by the golden answer.
\item Refusal Rate: The proportion of samples where the model refuses to answer the question.
\item VU/SU Disagreement Rate: The proportion of samples where SU and VU disagree, meaning one is above the threshold while the other is below. A lower disagreement rate suggests that the two uncertainties are well-calibrated.
% \item Absolute Difference: The average of absolute difference between normalized SU and VU. It provides a measure of how closely aligned the two types of uncertainties are across responses. A lower average absolute difference indicates better alignment.
\item Correlation Coefficient: The correlation coefficient between SU and VU measures the strength and direction of the linear relationship between these two uncertainties. A higher correlation coefficient indicates better alignment.
\item VU for Incorrect answer: The average of VU for incorrect responses. VU should be relatively high, indicating that the model is less confident in its incorrect outputs.
\item VU for Correct answer: The average of VU for correct responses. 
This serves as a reference metric to ensure that the VU for correct answers is relatively stable. 

\end{itemize}

\begin{table}[]
\centering
\begin{minipage}{0.48\linewidth}
\centering
\resizebox{\linewidth}{!}{%
\begin{tabular}{lccccc}
\toprule
\multicolumn{1}{c}{\textbf{Setting}} & \textbf{\begin{tabular}[c]{@{}c@{}}Conf. Hallu.\\ Rate ↓\end{tabular}} & \textbf{\begin{tabular}[c]{@{}c@{}}Disagree.\\ Rate ↓\end{tabular}} & \textbf{Corr.↑} & \textbf{\begin{tabular}[c]{@{}c@{}}VU for\\ Incorrect↑\end{tabular}} & \textbf{\begin{tabular}[c]{@{}c@{}}VU for\\ correct\end{tabular}} \\ \hline
\multicolumn{6}{c}{TriviaQA} \\ \hline
w/ calculated Us & \textbf{19.0} & 21.4 & 0.63 & \textbf{0.55} & 0.16 \\
w/ predicted Us & 22.3 & \textbf{13.5} & \textbf{0.86} & 0.49 & 0.20 \\ \hline
\multicolumn{6}{c}{NQ-Open} \\ \hline
w/ calculated Us & \textbf{26.2} & \textbf{18.9} & \textbf{0.69} & \textbf{0.54} & 0.24 \\
w/ predicted Us & 28.5 & 25.1 & 0.65 & 0.48 & 0.26 \\ \hline
\multicolumn{6}{c}{PopQA} \\ \hline
w/ calculated Us & \textbf{21.6} & 44.7 & 0.34 & \textbf{0.73} & 0.20 \\
w/ predicted Us & 29.7 & \textbf{42.3} & \textbf{0.41} & 0.59 & 0.39 \\ 
\bottomrule
\end{tabular}%
}
\caption{Ablation Study Results for Llama-3.1-8B-Instruct, showing the impact of replacing calculated uncertainty values with values predicted by probes on the hidden state of the last token of the question.}
\label{tab:ablation_mitigate_probe}
\end{minipage}
\hfill
\begin{minipage}{0.48\linewidth}
\centering
\resizebox{\linewidth}{!}{%
\begin{tabular}{lcccccc}
\toprule
\multicolumn{1}{c}{\textbf{Setting}} & \textbf{\begin{tabular}[c]{@{}c@{}}Conf. Hallu.\\ Rate ↓\end{tabular}} & \textbf{\begin{tabular}[c]{@{}c@{}}Disagree.\\ Rate ↓\end{tabular}} & \textbf{Corr.↑} & \textbf{\begin{tabular}[c]{@{}c@{}}VU for\\ Incorrect↑\end{tabular}} & \textbf{\begin{tabular}[c]{@{}c@{}}VU for\\ correct\end{tabular}} \\ \hline
\multicolumn{6}{c}{TriviaQA} \\ \hline
before & 23.3 & 21.5 & 0.59 & 0.5 & 0.16 \\
w/ Rand & 20.1 & 22.4 & 0.59 & 0.5 & 0.17 \\
w/ VUF & \textbf{19.0} & \textbf{21.4} & \textbf{0.63} & \textbf{0.55} & \textbf{0.16} \\ \hline
\multicolumn{6}{c}{NQ-Open} \\ \hline
before & 40.2 & 35.1 & 0.38 & 0.37 & 0.17 \\
w/ Rand & 35.2 & 26.7 & 0.45 & 0.38 & 0.17 \\
w/ TriviaQA VUF & \textbf{26.2} & 19.6 & \textbf{0.70} & \textbf{0.54} & \textbf{0.24} \\
w/ VUF & \textbf{26.2} & \textbf{18.9} & 0.69 & \textbf{0.54} & \textbf{0.24} \\ \hline
\multicolumn{6}{c}{PopQA} \\ \hline
before & 33.7 & 50.7 & 0.05 & 0.61 & 0.17 \\
w/ Rand & 28.8 & 47.6 & 0.16 & 0.63 & 0.18 \\
w/ TriviaQA VUF & 22.4 & \textbf{40.2} & \textbf{0.37} & 0.70 & 0.23 \\ 
w/ VUF & \textbf{21.6} & 44.7 & 0.34 & \textbf{0.73} & 0.20 \\
\bottomrule
\end{tabular}%
}
\caption{Ablation Study Results for Llama-3.1-8B-Instruct when the VUF from TriviaQA, NQ-Open, and PopQA datasets is replaced with: 
(1) VUF extracted from TriviaQA only, applied to intervene on NQ-Open and PopQA samples. 
(2) random vectors, applied to intervene on three datasets.}
\label{tab:ablation_mitigate_generalization_luf}
\end{minipage}
\end{table}

\paragraph{Result}
We compare results before and after the application of MUC with calculated uncertainties on TriviaQA, NQ-Open, and PopQA in Tab.~\ref{tab:mitigation}.
Overall, the application of MUC results in a substantial reduction of confident hallucinations at the cost of a small decrease in Correctness Rate.~\footnote{The correctness rate decreases when the correct answer exhibits high SU and the MUC intervention results in an abstention. This reduction stems from noise in measuring SU. For the same question, the sampled answers may contain diverse additional information, leading to a high calculated semantic entropy. Please see Appendix~\ref{app:decrease correctness} for an example.
Different trade offs between Hallucination Rate and Correct Rate can be obtained by varying the strength of the inference-time interventions. The optimal trade-off is highly application-dependent.}
% while maintaining a relatively stable correct rate.
The observed decrease in the VU/SU Disagreement Rate, along with an increase in the Correlation Coefficient, demonstrates improved calibration and alignment between verbal and semantic uncertainties across most models and datasets. 
While the VU for incorrect answers increased significantly, indicating reduced confidence in incorrect outputs, the VU for correct answers remained relatively unchanged after uncertainty calibration.\footnote{The increase in VU also for correct answers is not necessarily wrong, as it stems from the cases where the answer is correct, but the model is semantically uncertain about it.}
The fact that the trend is consistent across different models and sizes underscores the generality and effectiveness of our approach.
% We observe a similar phenomenon in \S~\ref{subsec: luf_causal_validation}, where Qwen2.5 is less sensitive to intervention.

% Our primary focus is on addressing confident hallucinations, but our approach also effectively reduces hallucinations.
% Our mitigation approach injects more VU, so that the output includes more uncertain answers and refusal responses. Consequently, we mitigate both types of hallucinations: confident hallucinations (incorrect and certain → uncertain/refusal) and conventional hallucinations (incorrect and non-refusal → refusal). 

As demonstrated in Tab.~\ref{tab:ablation_mitigate_probe}, mitigation results using probe-predicted uncertainties are somewhat worse but comparable to those based on calculated uncertainties. This suggests that probes can effectively predict uncertainties and mitigate hallucinations during inference.

To demonstrate the generality of our method across datasets, we applied VUFs derived from TriviaQA with calculated uncertainties to mitigate hallucinations in NQ-Open and PopQA. As illustrated in Tab.~\ref{tab:ablation_mitigate_generalization_luf}, the hallucination rate decreases, supporting the finding in \S~\ref{paragraph: luf consistent across datasets} that VUFs are consistent across different datasets. VUFs from one dataset can effectively control the VU level in other datasets.

To further prove the importance of using VUF in the MUC method, we perturb activations with random vectors with the same $\alpha$ and value range as the VUF, when the hallucination detector trigger. As shown in Tab.~\ref{tab:ablation_mitigate_generalization_luf}, applying random perturbations upon hallucination detection does lead to an improvement over the baseline. However, the results show that intervening on the VUF direction is significantly more effective than random perturbations.

This approach not only mitigates hallucinations but also enhances the overall reliability of the model's outputs.

\section{Related Work}
\label{sec:related_work}

We discuss relevant work on linear feature discovery and model steering in \S~\ref{subsec:semantic_space}. Here we present related work on other aspects of this work.

\subsection{Uncertainty in LLMs}
Uncertainty estimation has long been a cornerstone of reliable machine learning systems \citep{gawlikowski2023survey}. Recent success of LLMs in real-world applications have expanded the scope of uncertainty estimation research, prompting efforts to address unique challenges these models face in open-ended generation~\citep{huang2024survey,duan-etal-2024-shifting}.
One line of research  focuses on token-level uncertainty estimation over probabilities of LLM generations, such as predictive confidence or entropy. However, these approaches work at the level of tokens and do not capture the uncertainty over semantic meaning. Resampling-based methods address this limitation leveraging self-consistency across multiple generated responses \citep{duan-etal-2024-shifting, zhang-etal-2024-luq,farquhar2024detecting, wangself,malinin2020uncertainty, chen2024inside, gao-etal-2024-spuq}.
However, these methods do not fully account for the verbal uncertainty expressed by models. \citet{mielke-etal-2022-reducing} refers to a model’s verbalized expression of confidence as "linguistic confidence". This study involves manually annotating responses with labels of refusal, low or high confidence and training a specialized classifier on this dataset. Similarly, \citet{tomani2024uncertainty} introduces the concept of "in-dialogue uncertainty" by counting hedge words from a predefined list in model outputs.

\subsection{Hallucination Detection and Mitigation}
\paragraph{Detection}
% The detection of hallucinations in language models has garnered significant attention in recent research.
Studies have demonstrated that model uncertainty can serve as an indicator for identifying hallucinations~\citep{farquhar2024detecting,chen2024inside,zhang-etal-2023-enhancing-uncertainty,xiao2021hallucination}.
Other works have explored using the internal states of LLMs for detection~\citep{azaria2023internal,jillm,snyder2023early,kadavath2022language}. 
Additionally, some studies have focused on building annotated datasets and fine-tuning hallucination detectors on them~\citep{ji2024anah,Gu2024ANAHv2SA,mishra2024fine,li2023halueval,muhlgay2024generating,varshney2023stitch,yang2023new}. 
% \citet{zhang-etal-2023-enhancing-uncertainty} proposed a method that focuses on key aspects such as important keywords and unreliable tokens to enhance uncertainty-based detection. 
% Our work introduces a novel approach by integrating linguistic uncertainty with semantic uncertainty to enhance the accuracy and reliability of hallucination detection.
To the best of our knowledge, ours is the first work to show the effectiveness of combining VU and SU for hallucination detection.

\paragraph{Mitigation}
One approach to mitigating hallucinations is generating more faithful and factual answers include model editing~\citep{daheim2023elastic, ji-etal-2023-rho}, decoding rectification~\citep{rebuffel2022controlling, chuang2023dola, shi2023trusting,li2023inference}, mechanistic fine-tuning \citep{yu2024mechanistic,wu2024reft}, re-ranking~\citep{Gu2024ANAHv2SA} and variants of the Chain-of-Thought approach involving verification or reflection~\citep{dhuliawala2023chain,lei2023chain,ji2023towards,wang2023unleashing}. 
Alternative mitigation methods for improving the trustworthiness of models involve the use of abstention and controlled stopping mechanisms ~\citep{cheng2024can,duan-etal-2024-shifting, tomani2024uncertainty,feng-etal-2024-dont,zhang-etal-2024-r}. These works aim to completely refrain from answering the question when the model is uncertain, thereby reducing the likelihood of hallucinations. 
% \citet{ferrando2024know} propose to control hallucinations by manipulating directions in the representation space responsible for entity recognition.

Unlike abstention, which involves refusing to answer in the face of uncertainty, we aim to incorporate the uncertainty in the output text form. With similar motivation to this work, \citet{band2024linguistic} trains models to verbally convey the probability that their claims are correct; \citet{stengel2024lacie} fine-tunes the model based on user feedback regarding the perceived correctness of answers. Our work does not involve fine-tuning of the LLM, additional system prompt design, or sampling methods required by previous mitigation works.

% Our work does not include fine-tuning of the model, additional system prompt design, or sampling methods required by previous mitigation works.
% We showed that there exists universal LUFs that can be pre-comuputed in advance which saves on computation overhead. Our proposed method leverages the model's inherent ability to express nuanced uncertainty, allowing it to better manage and communicate its confidence levels. This approach not only mitigates hallucinations but also enhances the overall reliability of the model's outputs.  

% Uncertainty estimation (UQ) has long been a cornerstone of reliable machine learning systems, such as Bayesian approximation~\citep{gal2016dropout}, ensemble learning~\citep{hoffmann2021uncertainty}, and entropy measures~\citep{} have been widely explored~\citep{catak2024uncertainty,abdar2021review, namdari2019review}. 

% including SAR~\citep{duan-etal-2024-shifting}, LUQ~\citep{zhang-etal-2024-luq}, Semantic Entropy~\citep{farquhar2024detecting}, \citet{wangself,malinin2020uncertainty}, and EigenScore with hidden state~\citep{chen2024inside}. In addition, perturbation-based method SPUQ~\citep{gao-etal-2024-spuq} measure both aleatoric and epistemic uncertainties through strategic input variations coupled with output sampling.

% to prevent the generation of potentially inaccurate content

\section{Conclusion and Future Work}
\label{sec:conclusion}
We address the critical issue of hallucinations with overconfidence in LLMs.
% Our approach leverages recent advancements in understanding LLMs' internal states and semantic uncertainty to detect and mitigate this issue. 
We demonstrate that an underlying issue contributing to hallucinations is the misalignment between models' intrinsic semantic uncertainty and its expressed verbal uncertainty (VU). 
We discover the existence of a Verbal Uncertainty Feature (VUF), a single direction in the representation space that governs the VU. 
We leverage these insights in two applications:
(1) a hallucination detection method that integrates semantic and verbal uncertainties, outperforming methods relying solely on semantic uncertainty;
(2) a mitigation method, Mechanistic Uncertainty Calibration (MUC), based on aligning verbal uncertainty with the model’s semantic uncertainty by steering activations along the VUF direction during inference. 
Our findings suggest that LLMs can benefit from a more nuanced expression of uncertainty. Empirical results demonstrate a significant reduction in hallucinations and improved alignment, thereby enhancing the trustworthiness and reliability of LLM outputs.

For future work, one potential direction is to improve the generalizability of VUF across different LLM architectures and extend its application to long-form QA tasks. 
Additionally, investigating why activation steering sensitivity varies among LLM architectures could provide valuable insights. 
Finally, exploring how models represent uncertainty arising from various causes, such as underspecified questions, controversial topics, ethical dilemmas, and model ignorance, etc., would be an interesting next step in interpretability.

\section*{Limitations}
While our discovery of the Verbal and Semantic Uncertainty Framework (VUF) and the proposed method show promise in reducing hallucinations by calibrating uncertainty, there are several limitations to consider.
Firstly, our investigation is based on short-form (sentence-length) QA datasets, which may not fully capture the complexity of real-world scenarios.
Additionally, although we have demonstrated improvements in uncertainty calibration, the method's reliance on predefined probes and scores may limit its adaptability to unforeseen contexts or novel queries.
Lastly, our approach primarily focuses on enhancing the model's internal mechanisms for expressing uncertainty, which does not necessarily lead to correcting hallucinated answers.
Future work should address these aspects to develop a more comprehensive solution for mitigating hallucinations in LLMs.

\bibliographystyle{assets/plainnat}
\bibliography{paper}

\clearpage
\newpage
\beginappendix

\section{Prompts}
\label{app:prompts}

\subsection{Answer Generation Prompt}
\label{appendix:prompt_lu}
Prompt to generate an answer for the given question.
\begin{tcolorbox}[fontupper=\small, fontlower=\footnotesize]
\textbf{Answer Generation Prompt}
Please answer the following question.

Question: \{question\}

Answer:
\end{tcolorbox}

To calculate verbal uncertainty of question, we encourage LLM to express uncertainty.
\begin{tcolorbox}[fontupper=\small, fontlower=\footnotesize]
\textbf{Answer Generation Prompt for Verbal Uncertainty }

    Answer the following question using a succinct (at most one sentence) and full answer. If you are uncertain about your answer to the question, convey this uncertainty verbally by precisely hedging this answer. 
\end{tcolorbox}

\subsection{LLM-as-a-Judge Decisiveness Prompt}
\label{appendix:prompt_llm_judge}
Prompt to output a decisiveness score between 0 and 1 for the question and candidate answer. 
\begin{tcolorbox}[fontupper=\small, fontlower=\footnotesize]
\textbf{Verbal Uncertainty Judge Prompt}

You will be given a question and a proposed answer generated by a language model. 

Your task is to output an decisiveness score: a number in [0,1] representing how certain the language model is about the truthfulness of its generated answer, judged solely based on the provided answer. You should pay special attention to the usage of any hedging modifiers in the original answer, used to convey uncertainty in the truthfulness of the answer. A score close to 1.0 indicates that the model is very certain about the truthfulness answer.  And a score close to 0.0 means that the model is highly uncertain, and the answer likely contains some hedging expressions.  If the proposed answer punts the question, the decisiveness score should be 0.0.

Here are some examples: 

Question: criminal minds episode where jj becomes a profiler Proposed answer: I think JJ officially becomes a profiler in Criminal Minds season 6, episode 2. \\ Decisiveness score: 0.8

Question: criminal minds episode where jj becomes a profiler Proposed answer: JJ officially becomes a profiler in Criminal Minds season 6, in either the first or second episode. \\ Decisiveness score: 0.5

Question: criminal minds episode where jj becomes a profiler Proposed answer: I’m not really sure about this, but I think the episode in which JJ officially becomes a profiler in Criminal Minds may be episode 2 in season 6.\\  Decisiveness score: 0.6

Question: criminal minds episode where jj becomes a profiler Proposed answer: I don’t know which episode you’re referring to.\\  Decisiveness score: 0.0
\end{tcolorbox}

\subsection{LLM-as-a-Judge Accuracy Prompt}
\label{appendix:prompt_acc}
Prompt for LLM-as-a-judge of accuracy which asks to compare the golden answers and the predicted answer:

\begin{tcolorbox}[fontupper=\small, fontlower=\footnotesize]
\textbf{Prompt for Accuracy Judge}
We are assessing the quality of answers to the following question: \{question\}
The following are expected answers to this question: \{golden answers\}
The proposed answer is: \{predicted answer\}
Within the context of the question, does the proposed answer mean the same as any of the expected answers?
Respond only with yes or no.
Response:
\end{tcolorbox}

\section{Datasets}
\label{app:dataset}
To empirically demonstrate the VUFs explained above, we adopt three closed-book short-form QA datasets: TriviaQA~\citep{joshi2017triviaqa}, NQ-Open ~\citep{kwiatkowski2019natural}, and PopQA~\citep{mallen2023llm_memorization}.

TriviaQA~\footnote{\url{https://huggingface.co/datasets/mandarjoshi/trivia_qa}} consists of over 650,000 question-answer-evidence triples, including 95,000 question-answer pairs from trivia enthusiasts. Each question is supported by an average of six evidence documents. We use the RC version and sample 10,000 instances from the training set and 1,000 from the validation set for validation and 1,000 from the validation set for testing.

NQ-Open~\footnote{\url{https://huggingface.co/datasets/google-research-datasets/nq_open}} is an open-domain QA benchmark derived from Natural Questions, focusing on English Wikipedia content. We sampled 10,000 instances from the training set and 1,000 from the validation set for validation and testing.

PopQA~\footnote{\url{https://huggingface.co/datasets/akariasai/PopQA}} features 14,000 entity-centric QA pairs generated from Wikidata tuples. It includes annotations for subject and object entities, relationship types, and Wikipedia page views. We sampled 10,000 instances for training, 1,000 for validation, and 1,000 for testing.

\section{Experimental Details for Uncertainty Calculation}
\label{app:experimental_detail}
We adhere to the generation settings in the previous paper~\citep{kossen2024semantic,farquhar2024detecting} when calculating semantic uncertainty. We input a question into the language model and sample 10 sequences, using a temperature of 1 with nucleus sampling (P = 0.9) and top-K sampling (K = 50). Additionally, we generate a single sequence at a low temperature (0.1) to estimate the model's most likely answer to the query, which aids in assessing potential hallucinations. The generation process is conducted using a GPU H100.

\section{Miscalibration between Semantic and Verbal Uncertainties}
\label{app:examples_4cases}

Tab.~\ref{tab:refsual_ratio} show the proportion of four types of questions classified by the level of SU and VU.

\begin{table}[]
\centering
\begin{minipage}{0.48\linewidth}
\centering
\resizebox{\linewidth}{!}{
\begin{NiceTabular}{lcc|cc}
\toprule
\multicolumn{1}{c}{\multirow{2}{*}{\textbf{Dataset}}} & \multicolumn{2}{c|}{\textbf{Abstained}} & \multicolumn{2}{c}{\textbf{Complying}} \\
\multicolumn{1}{c}{} & \multicolumn{1}{c}{\textbf{\begin{tabular}[c]{@{}c@{}}Consistently \\ Abstained\end{tabular}}} & \multicolumn{1}{c|}{\textbf{\begin{tabular}[c]{@{}c@{}}Partly \\ Abstained\end{tabular}}} & \multicolumn{1}{c}{\textbf{Hallucinated}} & \multicolumn{1}{c}{\textbf{Correct}} \\
\hline
TriviaQA & 1.4 & 6.7 & 20.3 & 71.6 \\
NQ-Open & 3.2 & 9.0 & 28.3 & 59.5 \\
PopQA & 31.0 & 24.9 & 17.2 & 26.9 \\
\bottomrule
\end{NiceTabular}
}
\caption{Proportion of four types of responses: correct, hallucinated, partly abstained, and consistently abstained.}
\label{tab:refsual_ratio}
\end{minipage}
\hfill
\begin{minipage}{0.48\linewidth}
\centering
\resizebox{\linewidth}{!}{
\begin{NiceTabular}{crrcc}
\toprule
\textbf{Dataset} & \multicolumn{1}{c}{\textbf{\begin{tabular}[c]{@{}c@{}}Last Token\\ Hidden State\end{tabular}}} & \multicolumn{1}{c}{\textbf{Predicted Feature}} & \textbf{AUROC} & \textbf{ACC} \\ \hline
\multirow{6}{*}{\textbf{TriviaQA}} & \multirow{3}{*}{Question} & Semantic only & 66.85 & 66 \\
 &  & Verbal only & 68.48 & \textbf{70.9} \\
 &  & Combined & \textbf{-} & 70.4 \\ \cline{2-5} 
 & \multirow{3}{*}{Answer} & Semantic only & 74.03 & 70.9 \\
 &  & Verbal only & 68.61 & 69.8 \\
 &  & Combined & \textbf{-} & \textbf{74.3} \\ \hline
\multirow{6}{*}{\textbf{NQ-Open}} & \multirow{3}{*}{Question} & Semantic only & 54.07 & 53.9 \\
 &  & Verbal only & 50.9 & 58.5 \\
 &  & Combined & \textbf{-} & \textbf{74.7} \\ \cline{2-5} 
 & \multirow{3}{*}{Answer} & Semantic only & 61.32 & 57.4 \\
 &  & Verbal only & 50.64 & 61.2 \\
 &  & Combined & \textbf{-} & \textbf{79.1} \\ \hline
\multirow{6}{*}{\textbf{PopQA}} & \multirow{3}{*}{Question} & Semantic only & 70.17 & 65.6 \\
 &  & Verbal only & 35.96 & 43.4 \\
 &  & Combined & \textbf{-} & \textbf{75.8} \\ \cline{2-5} 
 & \multirow{3}{*}{Answer} & Semantic only & 69.91 & 67.8 \\
 &  & Verbal only & 34.21 & 39.4 \\
 &  & Combined & \textbf{-} & \textbf{77.9} \\
 \bottomrule
\end{NiceTabular}
}
\caption{Detection Results for Classifier Binarized Uncertainty Probe on Llama-3.1-8B-Instruct.}
\label{tab:SEP}
\end{minipage}
\end{table}

\footnotetext{Since we cannot get the combined probabilities of two uncertainties, we cannot get the AUROC score.}

Each example includes the following components:
\begin{itemize}
\item \textbf{Question:} The query or prompt being addressed.

\item \textbf{Answer to be Evaluated:} The most likely response under assessment for its level of hallucination.

\item \textbf{Golden Reference:} The correct and verified answer to the question.

\item \textbf{Semantic Entropy:} A measure of the variability or uncertainty in the responses.

\item \textbf{Sampled Responses:} A collection of ten responses, each marked with a \textit{semantic cluster index}. Responses sharing the same index belong to the same semantic group, indicating they have similar meanings or interpretations.
\end{itemize}

\subsection{Consistently Abstained Examples}
\begin{itemize}
  \setlength{\itemsep}{0pt} % Adjusts space between items
  \setlength{\parskip}{0pt} % Adjusts space between paragraphs
  \setlength{\parsep}{0pt}  % Adjusts space between paragraphs within an item
    \item \textbf{Question:}  Which TV creatures ate Blue String Pudding ?
    \item \textbf{Answer to be Evaluated:} I'm not aware of any information about TV creatures eating Blue String Pudding.
    \item \textbf{Golden Reference:}  ['The Soup Dragon', 'Clangers', 'Iron Chicken', 'The Iron Chicken', 'Soup Dragon', 'THE CLANGERS', 'The Clangers']
    \item \textbf{Semantic Entropy:} 0.00
    \item \textbf{Sampled Responses:}
    \begin{itemize}
       \item 0 I am sorry, but I don't have information about TV creatures eating Blue String Pudding.

        \item 0 I'm not aware of any information about TV creatures eating blue string pudding. This may be a specific reference
        
        \item 0 I don't have information about the TV creatures eating Blue String Pudding.
        
        \item 0 I'm not aware of any specific TV show that features a scene where a creature eats Blue String Pudd
        
        \item 0 I don't have information about the TV creatures that ate Blue String Pudding. Can you provide more
        
        \item 0 I am sorry, but I do not have information about TV creatures that ate Blue String Pudding.
        
        \item 0 I am sorry, I do not have information on TV creatures eating blue string pudding. Can I help you
        
        \item 0 I'm not aware of the specific context or show you're referring to.
        
        \item 0 I'm sorry, but I am not aware of a specific answer that relates to a type of TV creature
        
        \item 0 I do not have information on which TV creatures ate Blue String Pudding.
    \end{itemize}
\end{itemize}

\subsection{Partly Abstained Examples}
\begin{itemize}
    \item \textbf{Question:} The opera The Tsar Sultan contains what famous musical piece?
    \item \textbf{Answer to be Evaluated:} The opera "The Tsar Sultan" is a work by Nikolai Rimsky-Korsakov. However, I am unable to verify if it contains a famous musical piece.
    \item \textbf{Golden Reference:} ['The Flight of the Bumble Bee', 'The flight of the bumble bee', 'Flight of the Bumble-bee', 'The flight of the bumblebee', 'The Flight of the Bumblebee', 'Flight of the Bumblebee (Rimsky-Korsakov)', 'Flight of the Bumble Bee', 'Flight of the bumble bee', 'Flight of the bumblebee', 'Flight of the Bumblebee']
    \item \textbf{Semantic Entropy:} 2.30
    \item \textbf{Sampled Responses:}
    \begin{itemize}
       \item 0 The opera "The Tsar Sultan" is by Nikolai Rimsky-Korsakov. "Song of the Valkyries" or "Song of the Dutchman" are not well known famous musical pieces by this opera.

\item 1 The Tsar Sultan is an opera composed by Nikolai Rimsky-Korsakov. It contains the famous musical piece "Song of the Invisible One" and "Song of the Fisherman," but the most well-known piece in this opera is (hallucinated)

\item 2 The Tsar Sultan is an opera by Nikanor Villamil, composed in 1836.

\item 3 The famous musical piece from the opera 'The Tsar Sultan' is the famous aria "Kalmusinkoy". (hallucinated)

% \item 4 The opera "The Tsar Sultan" (Italian: L'Imperatore di Lagos; Russian: Цар-молочник) was written by Niccolò Jommelli.
\item 4 The opera "The Tsar Sultan" was written by Niccolo Jommelli.

\item 5 The Tsar Sultan, an opera by Giuseppe Verdi, contains the famous musical piece "Song of the Flea" however it could be that you are referring to the 1862 French opera "The Tsar and the False (hallucinated)

\item 6 The opera "The Tsar Sultan" contains the famous musical piece "I Puritani" is not correct, I believe you are referring to the Opera "The Tsar Sultan" actually refers to the opera "Il Turco in Italia

\item 7 The Tsar Sultan is a 19th-century French opera by Niccolò Paganini's student and composer, Michele Carrara. However, I couldn't find information about a famous piece from The Tsar Sultan opera.

\item 8 The opera "The Tsar Sultan" contains the famous musical piece "Carmen Suite" is not correct but it does contain a famous piece called "Song of the Dutch Sailors".(hallucinated)

\item 9 The opera "The Tsar Sultan" by Nikolai Rimsky-Korsakov contains the famous musical piece "Flight of the Bumblebee". (correct)
    \end{itemize}
\end{itemize}

\begin{itemize}
    \item \textbf{Question:} When did McDonald's sell 1 million burgers?
    \item \textbf{Answer to be Evaluated:} I am unable to verify when McDonald's sold 1 million burgers.
    \item \textbf{Golden Reference:} ['By 1965']
    \item \textbf{Semantic Entropy:} 1.42
    \item \textbf{Sampled Responses:}
    \begin{itemize}
        \item 0 McDonald's sold its one millionth burger in 1955 (hallucinated)

\item 1 I'm sorry, but I am unable to verify when McDonald's sold

\item 2 However, I need more information about which McDonald's you are referring to.

\item 0 McDonald's first sold 1 million burgers in 1954 (hallucinated)

\item 3 McDonald's is a fast-food chain that has been selling burgers for many (hallucinated)

\item 1 I am unable to verify when McDonald's sold 1 million burgers.

\item 4 McDonald's was founded by brothers Richard and Maurice McDonald and initially operated a

\item 1 I'm not aware of a specific date when McDonald's sold 1 million

\item 0 McDonald's achieved this milestone on April 15, 1955 (hallucinated)

\item 1 I am unable to verify when McDonald's first sold 1 million burgers.
    \end{itemize}
\end{itemize}

\subsection{``Hallucinations'' with Low SU}
\label{app:low_SU_hallucinations}
Some samples with low SU are labeled as hallucinations and can be categorized into several cases:

(1) The labeling is correct:
\vspace{-0.5pt}
\begin{itemize}
 \setlength{\itemsep}{0pt} % Adjusts space between items
  \setlength{\parskip}{0pt} % Adjusts space between paragraphs
  \setlength{\parsep}{0pt}  % Adjusts space between paragraphs within an item
\item The generated content is consistently hallucinated (Example 0)
\item The answer being evaluated is hallucinated (temperature=0.1), but the sampled responses are correct (temperature=1.0). (Example 1)
\end{itemize}

(2) The labeling is wrong:
\begin{itemize}
 \setlength{\itemsep}{0pt} % Adjusts space between items
  \setlength{\parskip}{0pt} % Adjusts space between paragraphs
  \setlength{\parsep}{0pt}  % Adjusts space between paragraphs within an item
\item The golden answer itself is incorrect in the dataset. (Example 2)
\item The generated answer is similar, but not identical, to the correct answer. (Example 3)
\item The generated answer matches the correct answer, but the language model's judgment is incorrect. (Example 4)
\end{itemize}
\paragraph{Example 0 - Consistent Hallucination}
\begin{itemize}

  \setlength{\itemsep}{0pt} % Adjusts space between items

  \setlength{\parskip}{0pt} % Adjusts space between paragraphs

  \setlength{\parsep}{0pt}  % Adjusts space between paragraphs within an item

    \item \textbf{Question:} What is a third divided by a half?

    \item \textbf{Answer to be Evaluated:} 1/2 (half) = 0.5 

3/0.5 = 6

    \item \textbf{Golden Reference:} ['Two-thirds']

    \item \textbf{Semantic Entropy:} 0.0

    \item \textbf{Sampled Responses:}

    \begin{itemize}

       \item 0 To find the answer, we need to divide 3 by 0.5.

3 ÷ 0.5 = 6

So the answer is 6.

        \item 0 To find the answer, we need to divide 3 by 0.5 (half). 

3 ÷ 0.5 = 6

        \item 0 To solve this, we need to perform the calculation: 

3 (third) / 0.5 (half)

Since a third is 1/3 and a half is

        \item 0 To find the answer, we need to perform the following calculation: 

3 (third) / 0.5 (half) 

3 / 0.5 = 6

        \item 0 To find the answer, you need to divide 3 by 0.5 (half). 

3 / 0.5 = 6 

So, the answer is

        \item 0 To find the answer, we need to divide 3 (the third) by 0.5 (the half). 

3 ÷ 0.5 = 6

        \item 0 To find the answer, we need to divide a third by a half. 

A third is 1/3, and a half is 1/2. 

To divide

        \item 0 To find the answer, we need to perform the division.

1/2 (a half) = 0.5

3 (a third) ÷ 0.5

        \item 0 To find the answer, divide 3 by 0.5. 

3 ÷ 0.5 = 6

        \item 0 To solve this problem, we need to follow the order of operations (PEMDAS):

1. Divide a half (1/2) into the fraction 3:

    \end{itemize}

\end{itemize}

\paragraph{Example 1 - Hallucination, but other sampled responses are consistently correct.}
\begin{itemize}

  \setlength{\itemsep}{0pt} % Adjusts space between items

  \setlength{\parskip}{0pt} % Adjusts space between paragraphs

  \setlength{\parsep}{0pt}  % Adjusts space between paragraphs within an item

    \item \textbf{Question:} The first synthetic organic chemical dye, mauveine, was discovered in 1856 as a byproduct of the search for the synthetic form of what?

    \item \textbf{Answer to be Evaluated:} Quinoline.

    \item \textbf{Golden Reference:} ['Chinin', 'Quinine', 'Chinine', 'Kinine', 'Qualaquin', 'Coco-Quinine', 'Aflukin', 'Quinnine', 'Quinine powder', 'ATC code P01BC01', 'Legatrin']

    \item \textbf{Semantic Entropy:} 0.0

    \item \textbf{Sampled Responses:}

    \begin{itemize}

       \item 0 Quinine. The first synthetic organic chemical dye, mauveine, was discovered by William Perkin in 1856 as a byproduct of his search for the synthetic form of quinine.

        \item 0 The answer is quinine. Mauveine was discovered by William Henry Perkin in 1856, and it was a byproduct of his attempts to synthesize quinine

        \item 0 The answer to the question is quinine. The first synthetic organic chemical dye, mauveine, was discovered in 1856 as a byproduct of the search for the synthetic form of quinine

        \item 0 Quinine.

        \item 0 The answer is quinine. Mauveine was discovered in 1856 by William Henry Perkin, who was attempting to synthesize quinine.

        \item 0 The answer is quinine. Mauveine, the first synthetic organic chemical dye, was discovered in 1856 by William Henry Perkin as a byproduct of his search for the synthetic form

        \item 0 The first synthetic organic chemical dye, mauveine, was discovered in 1856 as a byproduct of the search for the synthetic form of quinine.

        \item 0 Quinine. The discovery of mauveine, a synthetic organic chemical dye, was a byproduct of the search for a synthetic form of quinine in the 1850s by French chemist

        \item 0 The first synthetic organic chemical dye, mauveine, was discovered in 1856 as a byproduct of the search for the synthetic form of quinine.

        \item 0 The answer to the question is Quinine.

    \end{itemize}

\end{itemize}

\paragraph{Example 2 - Wrong Golden Reference}

\begin{itemize}

  \setlength{\itemsep}{0pt} % Adjusts space between items

  \setlength{\parskip}{0pt} % Adjusts space between paragraphs

  \setlength{\parsep}{0pt}  % Adjusts space between paragraphs within an item

    \item \textbf{Question:} The scientific term fission means?

    \item \textbf{Answer to be Evaluated:} The scientific term "fission" means the process of splitting or dividing a nucleus of an atom into two or more smaller nuclei, often accompanied by the release of a large amount of energy.

    \item \textbf{Golden Reference:} ['Split (album)', 'Split (disambiguation)', 'Split (song)', 'Split', 'Split (single)']

    \item \textbf{Semantic Entropy:} 0.0

    \item \textbf{Sampled Responses:}

    \begin{itemize}

       \item 0 The scientific term 'fission' means the process by which an atomic nucleus splits into two or more smaller nuclei, along with the release of a large amount of energy.

        \item 0 The scientific term "fission" refers to the process by which an atomic nucleus splits into two or more smaller nuclei, along with the release of a large amount of energy.

        \item 0 The scientific term "fission" refers to the process of splitting or division of a nucleus of an atom into two or more smaller nuclei, often accompanied by the release of energy

        \item 0 The scientific term 'fission' refers to a process in which an atomic nucleus splits into two or more smaller nuclei, along with the release of a significant amount of energy.

        \item 0 The scientific term "fission" means the process of splitting or dividing into two or more parts, often used to describe the splitting of an atom or a cell.

        \item 0 The scientific term "fission" means the process of splitting or dividing something, typically into two or more smaller parts. This term is commonly used in various scientific fields, such

        \item 0 The scientific term "fission" refers to a process in which an atomic nucleus splits into two or more smaller nuclei, along with the release of a significant amount of energy.

        \item 0 The scientific term "fission" refers to a type of nuclear reaction where an atomic nucleus splits into two or more smaller nuclei, often accompanied by the release of a large amount

        \item 0 The scientific term 'fission' refers to the process by which an atomic nucleus splits into two or more smaller nuclei, often releasing a significant amount of energy in the process.

        \item 0 The scientific term "fission" means the process by which an atomic nucleus splits into two or more smaller nuclei, releasing a large amount of energy in the process.

    \end{itemize}

\end{itemize}

\paragraph{Example 3 - Incomplete Golden Reference}
% \begin{itemize}

%   \setlength{\itemsep}{0pt} % Adjusts space between items

%   \setlength{\parskip}{0pt} % Adjusts space between paragraphs

%   \setlength{\parsep}{0pt}  % Adjusts space between paragraphs within an item

%     \item \textbf{Question:} A ewer is a large?

%     \item \textbf{Answer to be Evaluated:} A ewer is a large vase or pitcher, typically used for holding and pouring water.

%     \item \textbf{Golden Reference:} ['Toby jugs', 'Toby Jugs', 'Jug', 'Character jug', 'Jug (container)', 'Jugs']

%     \item \textbf{Semantic Entropy:} 0.0

%     \item \textbf{Sampled Responses:}

%     \begin{itemize}

%        \item 0 A ewer is a large vase or jug.

%         \item 0 A ewer is a large pitcher or vase, typically

%         \item 0 A ewer is a large jug or vase used for

%         \item 0 A ewer is a large jug or vase, typically

%         \item 0 A ewer is a large vase or container.

%         \item 0 A ewer is a large container, typically made of

%         \item 0 A ewer is a large pitcher or vase used to

%         \item 0 A ewer is a large jug or a container for

%         \item 0 A ewer is a large vessel.

%         \item 0 A ewer is a large vessel.

%     \end{itemize}

% \end{itemize}
\begin{itemize}

  \setlength{\itemsep}{0pt} % Adjusts space between items

  \setlength{\parskip}{0pt} % Adjusts space between paragraphs

  \setlength{\parsep}{0pt}  % Adjusts space between paragraphs within an item

    \item \textbf{Question:} Plovers or dotterels are a type of birds classed as a?

    \item \textbf{Answer to be Evaluated:} Plovers or dotterels are a type of birds classed as Charadriiformes.

    \item \textbf{Golden Reference:} ['Waders', 'Waders (footwear)', 'Fishing wader']

    \item \textbf{Semantic Entropy:} 0.0

    \item \textbf{Sampled Responses:}

    \begin{itemize}

       \item 0 Charadriiformes is the answer. Charadriiformes is an order of small to medium-sized wading birds that includes plovers, dotterels, oystercatchers, avocets, stilts, and lapwings, among others.

        \item 0 Plovers or dotterels are a type of birds classed as Charadriiformes.

        \item 0 The answer to the question is "wading birds" or more specifically "shorebirds". However, in taxonomy, Plovers or dotterels are classed as "charadriiform birds" or more specifically "charadriids" or "plover-like shorebirds".

        \item 0 Plovers or dotterels are a type of birds classed as a shorebird.

        \item 0 Plovers or dotterels are a type of birds classed as Charadriiformes.

        \item 0 Plovers or dotterels are a type of birds classed as Charadriiformes, but more specifically they belong to the family Charadriidae.

        \item 0 Shorebirds.

        \item 0 Plovers or dotterels are a type of birds classed as shorebirds.

        \item 0 The answer is Shorebirds.

        \item 0 Plovers or dotterels are a type of birds classed as Charadriiformes.

    \end{itemize}

\end{itemize}
Explanation: Waders are part of the Charadriiformes order and include families such as Scolopacidae (sandpipers, snipes) and Charadriidae (plovers, lapwings).
\paragraph{Example 4 - Judge LLM error}
\begin{itemize}

  \setlength{\itemsep}{0pt} % Adjusts space between items

  \setlength{\parskip}{0pt} % Adjusts space between paragraphs

  \setlength{\parsep}{0pt}  % Adjusts space between paragraphs within an item

    \item \textbf{Question:} What is the name of the yacht that won the Royal Yacht Squadron's regatta around the Isle of Wight in 1851 and lent its name to a famous sports prize?

    \item \textbf{Answer to be Evaluated:} The yacht that won the Royal Yacht Squadron's regatta around the Isle of Wight in 1851 and lent its name to a famous sports prize is the "America".

    \item \textbf{Golden Reference:} ['The United States of America', 'United States Of Amerca', ...]

    \item \textbf{Semantic Entropy:} 0.0

    \item \textbf{Sampled Responses:}

    \begin{itemize}

       \item 0 The yacht that won the Royal Yacht Squadron's regatta around the Isle of Wight in 1851 and lent its name to a famous sports prize is the 'America' yacht.

        \item 0 The yacht that won the Royal Yacht Squadron's regatta around the Isle of Wight in 1851 and lent its name to a famous sports prize is "America". The America's Cup

        \item 0 The yacht that won the Royal Yacht Squadron's regatta around the Isle of Wight in 1851 and lent its name to a famous sports prize is "America."

        \item 0 The yacht that won the Royal Yacht Squadron's regatta around the Isle of Wight in 1851 and lent its name to a famous sports prize is "America".

        \item 0 The yacht you are referring to is the 'America'. It won the Royal Yacht Squadron's regatta around the Isle of Wight in 1851, and its victory led to the establishment

        \item 0 The answer is 'America'.

        \item 0 The answer is the "America". The yacht 'America' was the first to successfully complete the Royal Yacht Squadron's regatta around the Isle of Wight in 1851.

        \item 0 The yacht you're referring to is "America".

        \item 0 The yacht that won the Royal Yacht Squadron's regatta around the Isle of Wight in 1851 and lent its name to a famous sports prize is the America.

        \item 0 The yacht that won the Royal Yacht Squadron's regatta around the Isle of Wight in 1851 and lent its name to a famous sports prize is 'America'.

    \end{itemize}

\end{itemize}

\section{Verbal Uncertainty Feature (VUF)}
\label{app:vuf}

\subsection{Selected Layers for VUF}
Based on the visualization and consistency across different datasets discussed in \S~\ref{subsec: discovery of vuf}, we have selected the following layers for each model:
\begin{itemize}
\item  Llama-3.1-8B-Instruct: Layers 15 to 31
\item  Mistral-7B-Instruct-v0.3: Layers 15 to 31
\item  Qwen2.5-7B-Instruct: Layers 16 to 27
\end{itemize}

% computing embedding similarities with expressions of subjective uncertainty (e.g. "I don't know") and universal uncertainty (e.g. "It is not known").
% \textbf{(2) Embedding Similarity with Uncertainty Expressions}
% In addition to LLM-as-a-Judge method we experiment with alternatives: embedding similarities with uncertainty expressions. We generated short lists of expressions of subjective and universal uncertainty, which we donted as \textit{esu} and \textit{euu}. We use NV-Embed-v2 \citep{lee2024nv}, a generalist embedding model, to embed the generated answers and two types of uncertainty expressions separately. Then we calculate the cosine similarity between them as verbal uncertainty scores. 

\begin{figure*}[ht]
    \centering
    \includegraphics[width=\textwidth]{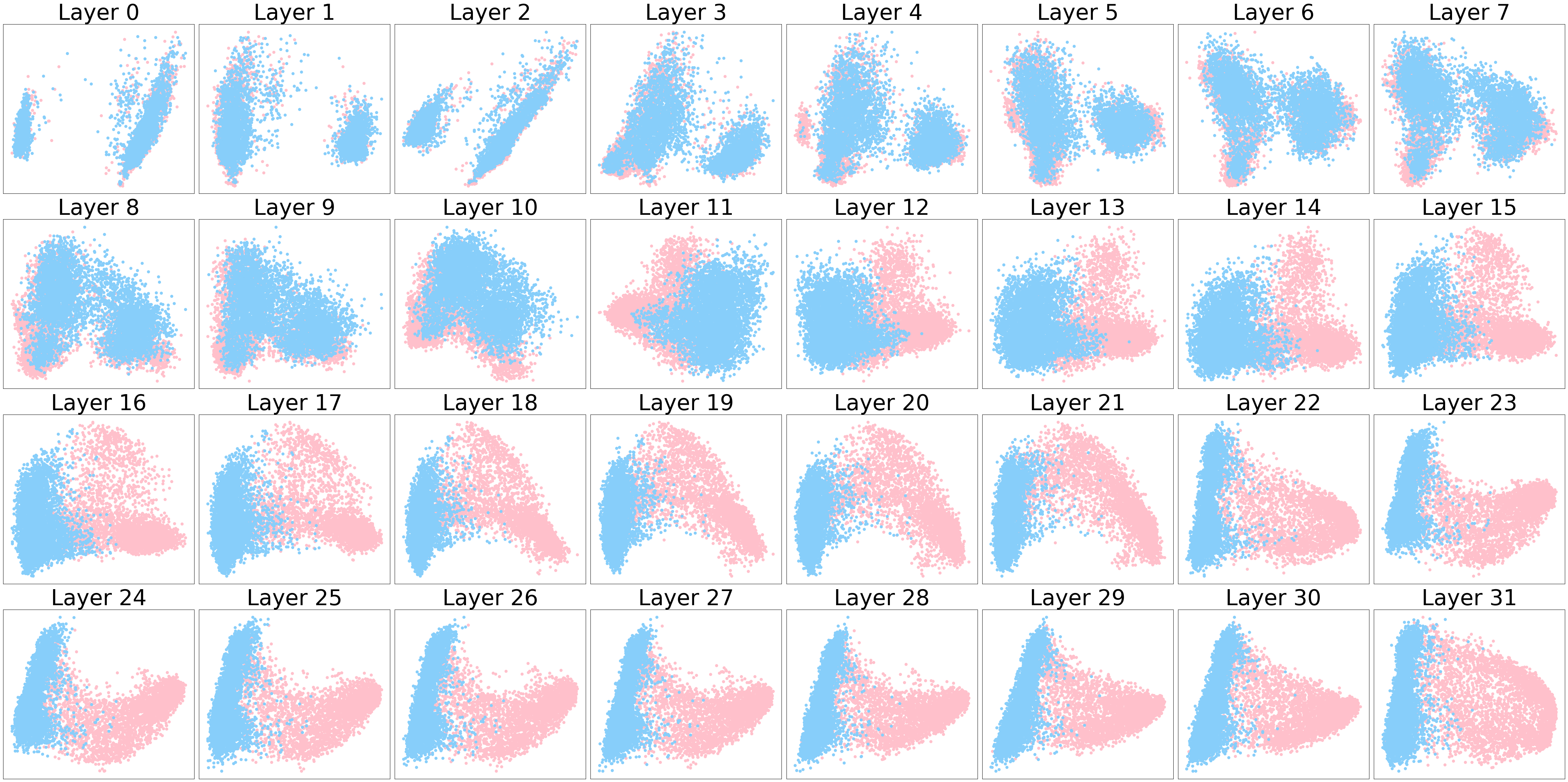}
    \caption{Visualization of verbalized certain vs. uncertain query representations from Llama-3.1-8B-Instruct for three datasets: TriviaQA, NQ-Open, and PopQA.}
    \label{fig:linear_lu_llama}
\end{figure*}

\begin{figure*}[ht]
    \centering
    \includegraphics[width=\textwidth]{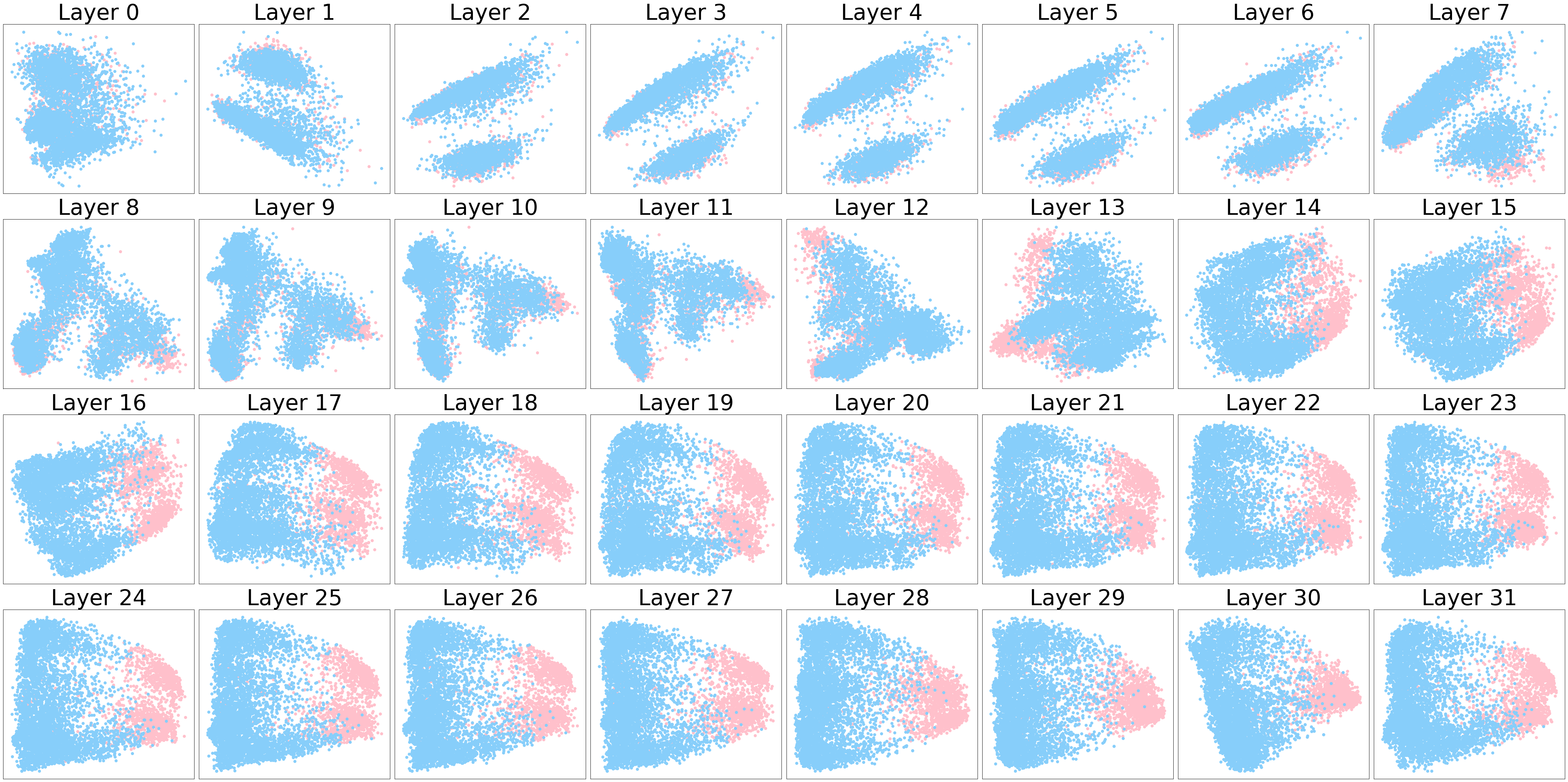}
    \caption{Visualization of verbalized certain vs. uncertain query representations from Mistral-7B-Instruct-v0.3 for three datasets: TriviaQA, NQ-Open, and PopQA.}
    \label{fig:linear_lu_mistral}
\end{figure*}

\begin{figure*}[ht]
    \centering
    \includegraphics[width=\textwidth]{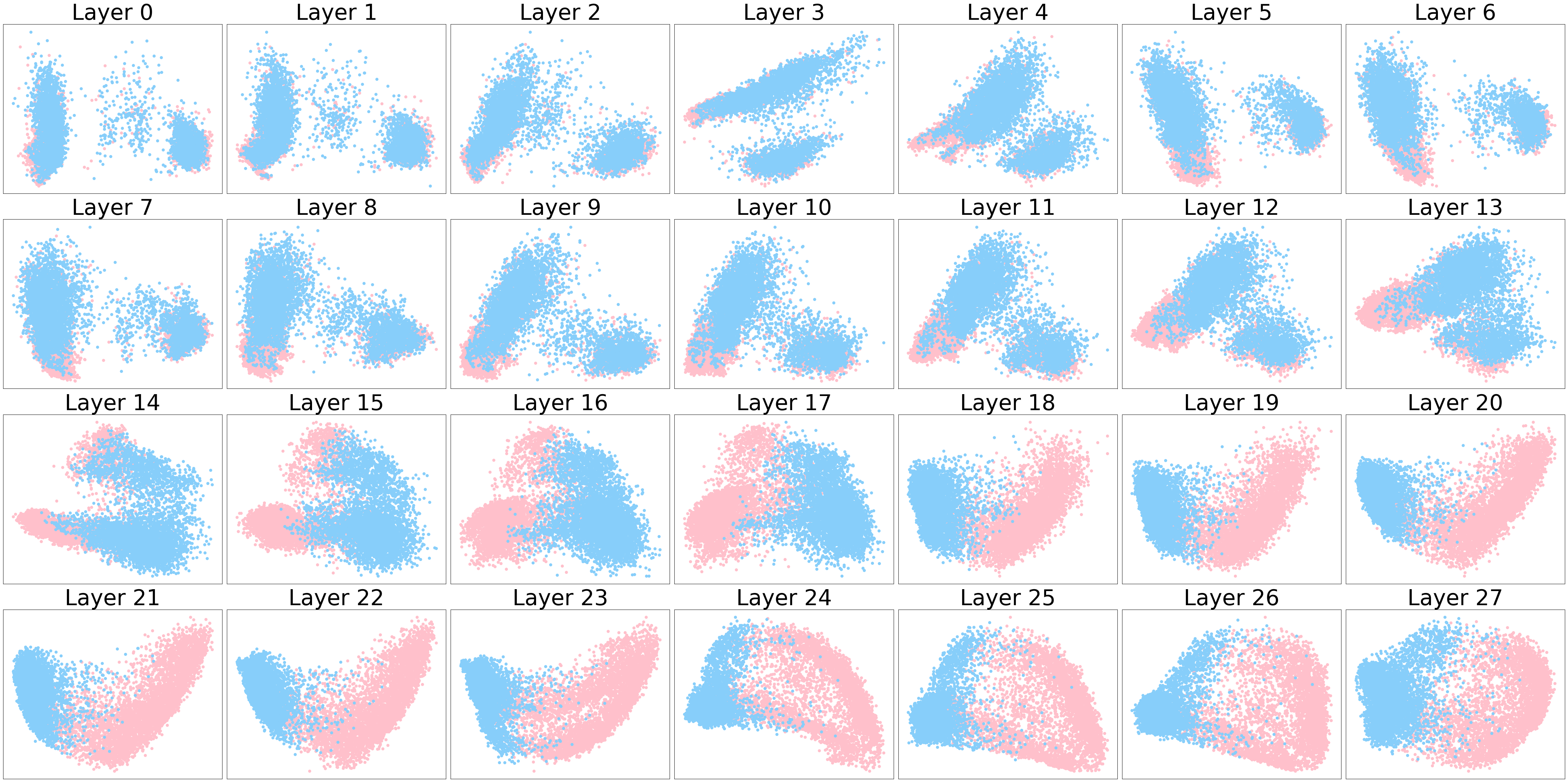}
    \caption{Visualization of verbalized certain vs. uncertain query representations from Qwen2.5-7B-Instruct for three datasets: TriviaQA, NQ-Open, and PopQA.}
    \label{fig:linear_lu_qwen}
\end{figure*}

%%%%%%%%%%%%%%%%%%%%%%%%%%%%%%%%%%%%%%%%%%%%%%%%%%%%%
%%%%%%%%%%%%%%%%%%%%%%%%%%%%%%%%%%%%%%%%%%%%%%%%%%%%%
\subsection{Cosine Similarity between VUFs from different verbal uncertainty scores.}
\label{app:esu_euu}
\begin{figure*}[ht]
    \centering
    \includegraphics[width=\textwidth]{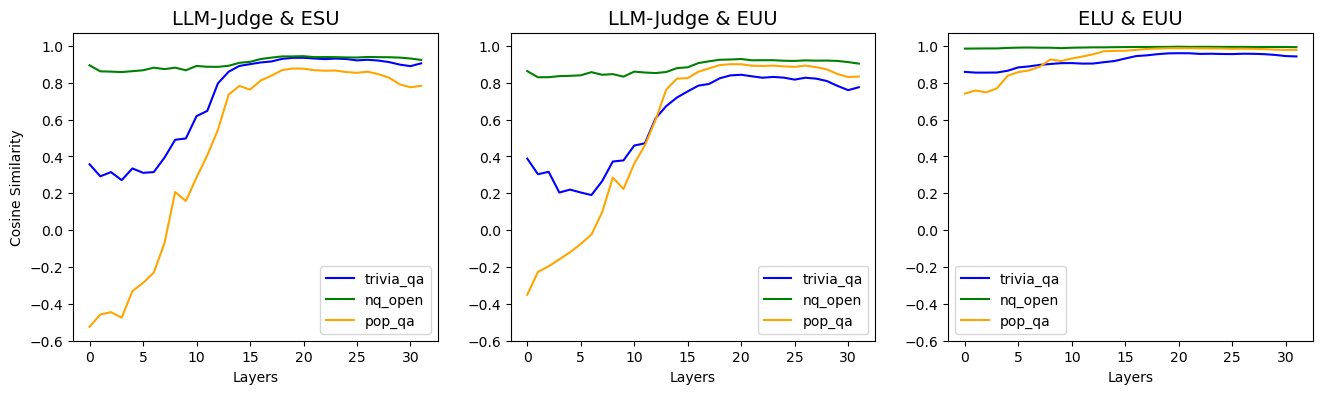}
    \caption{Cosine Similarity between VUFs from different VU scores on different datasets for Llama-3.1-8B-Instruct model.}
    \label{fig:cos_sim_lu_score_trivia_llama}
\end{figure*}

In addition to LLM-as-a-Judge method outlined in \S~\ref{subsec:verbal_u} we experiment with alternatives: embedding similarities with uncertainty expressions. We generated short lists of  expressions of subjective uncertainty (e.g. "I don’t know") and universal uncertainty (e.g. "It is not known"), denoted as ESU and EUU scores. We use NV-Embed-v2 \citep{lee2024nv}, a generalist embedding model, to embed the generated answers and two types of uncertainty expressions separately.

To compare each verbal uncertainty score out of LLM-Judge, ESU, and EUU, we construct $\mathcal{D}_{uncertain}$ and $\mathcal{D}_{certain}$. using each method. We then follow the steps outlined in \S~\ref{subsec:luf_feature_extract} and calculate the VUFs as described in Equation \ref{equation: luf}. We run our experiments on each of three datasets separately using Llama-3.1-8B-Instruct model. Figure \ref{fig:cos_sim_lu_score_trivia_llama} illustrates the cosine similarity of VUFs from each layer of examples obtained with different VU scores. We observe a high correlation between the three different scores for VUFs in the middle and subsequent layers.  These results demonstrate that our observations are consistent regardless of the choice of verbal uncertainty score.
\paragraph{Prototypical Expressions of Subjective Uncertainty (ESU)}
\begin{itemize}
  \setlength{\itemsep}{0pt} 
  \setlength{\parskip}{0pt}
  \setlength{\parsep}{0pt} 
    \item I'm not entirely sure, but...
    \item That's a tough one; let me think for a moment.
    \item I'd have to double-check on that.
    \item My answer might not be entirely accurate, but...
    \item I'm still considering the possibilities.
    \item I'm not confident in my answer, but I'll give it a shot.
    \item This is just an educated guess, but...
    \item I've heard conflicting information on this topic.
    \item My knowledge on this subject is limited.
    \item I'm not up-to-date on the latest developments.
    \item I'm starting to get out of my depth here.
    \item This is a bit beyond my expertise.
    \item I'm not familiar with that specific aspect.
    \item My understanding is incomplete.
    \item I'd need more context to provide a better answer.
    \item I'm really not sure about this one.
    \item My answer would be purely speculative.
    \item I've never encountered this situation before.
    \item I'm not aware of any definitive answer.
    \item The data on this topic is inconclusive.
    \item To be honest, I'm stumped.
    \item I'm having trouble finding a clear answer.
    \item My response would be a wild guess.
    \item I'm completely out of my element here.
    \item I wouldn't want to hazard a guess.
    \item Your guess is as good as mine.
    \item I wouldn't even venture a guess.
    \item It's impossible for me to say.
    \item There's too much ambiguity to provide an answer.
    \item I'm at a complete loss.
    \item I simply don't know.
    \item No idea, sorry.
    \item Not a clue.
    \item I'm clueless on this one.
    \item No answer comes to mind.
    \item That's outside my area of expertise.
    \item I'd rather not speculate.
    \item More research is needed to answer that.
    \item I'm still learning about this topic.
    \item There's no clear consensus on this issue.
    \item My answer would be unreliable.
    \item I wouldn't trust my own judgment on this.
    \item I've got nothing concrete to offer.
    \item No clear answer presents itself.
    \item I'd rather defer to someone else's expertise.
    \item I'm uncertain and unwilling to guess.
    \item Too many variables make it hard to answer.
    \item I lack sufficient information to respond.
    \item Any answer I gave would be unsatisfactory.
    \item Frankly, I'm baffled.
\end{itemize}

\paragraph{Prototypical Expressions of Universal Uncertainty (EUU)}
\begin{itemize}
  \setlength{\itemsep}{0pt} 
  \setlength{\parskip}{0pt}
  \setlength{\parsep}{0pt} 
    \item I'm not entirely sure about this.
    \item The answer is unclear at this time.
    \item More research is needed to determine the answer.
    \item This is still an open question.
    \item There's ongoing debate about this topic.
    \item It's difficult to say for certain.
    \item I couldn't find any reliable sources on this.
    \item The information available is limited.
    \item We don't have enough data to make a conclusion.
    \item This is a complex issue with no easy answer.
    \item I'm not aware of any definitive answer.
    \item The answer may depend on various factors.
    \item This is a topic of ongoing investigation.
    \item There's no straightforward answer to this question.
    \item Different perspectives offer varying insights.
    \item The situation is more nuanced than it seems.
    \item We need more context to provide an accurate answer.
    \item The answer might be subjective and dependent on interpretation.
    \item There's no clear consensus on this matter.
    \item Further analysis is required to determine the answer.
    \item Unfortunately, we can't provide a definitive answer.
    \item The question is too broad to give a specific answer.
    \item There are many variables at play here.
    \item We're dealing with incomplete information.
    \item The answer could go either way, depending on assumptions.
    \item This is a highly speculative area of inquiry.
    \item We're venturing into uncharted territory here.
    \item The data is inconclusive, and more study is needed.
    \item There's significant disagreement among experts.
    \item No clear pattern or trend emerges from the data.
    \item Honestly, we just don't know yet.
    \item The answer remains elusive despite our best efforts.
    \item This is a mystery waiting to be solved.
    \item We're stumped – more investigation is required.
    \item There's too much uncertainty to give a confident answer.
    \item Our current understanding is insufficient to answer this question.
    \item We're pushing the boundaries of human knowledge here.
    \item The question itself is still being refined.
    \item A definitive answer may never be possible.
    \item We're in unexplored territory, and caution is advised.
    \item Could you rephrase the question? It's unclear what you're asking.
    \item I'm having trouble understanding the context of your question.
    \item This question appears to be based on a false assumption.
    \item The question is too vague to provide a meaningful answer.
    \item We need to clarify some terms before proceeding.
    \item The question seems to be self-contradictory.
    \item I think there may be a misunderstanding here.
    \item Could you provide more background information on this question?
    \item This question doesn't seem to make sense in the given context.
    \item Nobody knows, and it's unlikely we'll ever find out (the ultimate cop-out!)
    \item Nobody knows.
    \item This question does not make any sense.
    \item That's an impossible question to answer.
\end{itemize}

\subsection{Cosine Similarity between VUFs from different LLM-as-a-Judge models.}
\label{app:llm_judge_sim}
\begin{figure*}[ht]
    \centering
    \includegraphics[width=\textwidth]{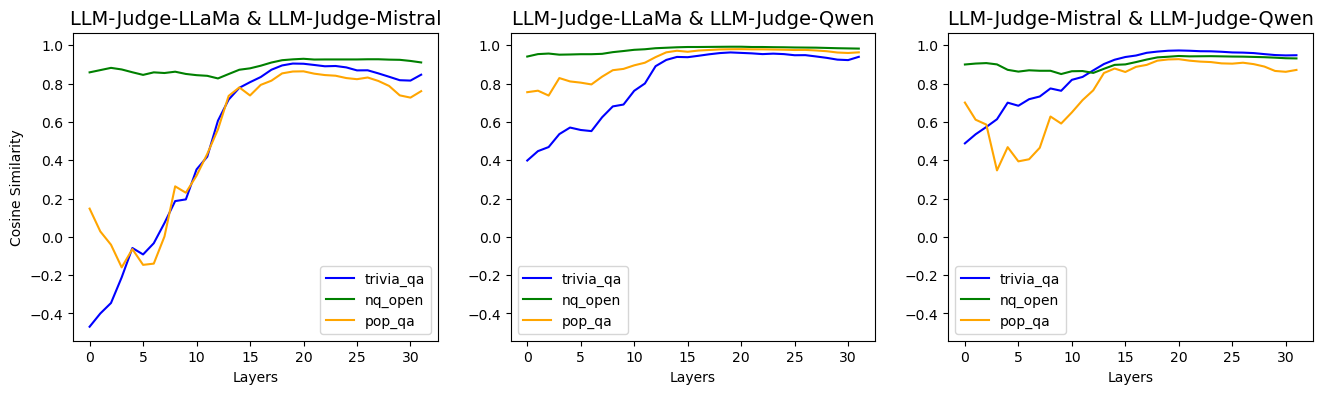}
    \caption{Cosine Similarity between VUFs from VU scores using different LLM-as-a-Judge models on different datasets for Llama-3.1-8B-Instruct model.}
    \label{fig:cos_sim_lu_score_llm_judge}
\end{figure*}

Continuing the discussion on LLM-as-a-Judge method for quantifying VU, we experiment with alternatives: use Mixtral-8x7B-Instruct-v0.1 and Qwen2.5-72B-Instruct as a LLM-as-a-Judge models.
Figure \ref{fig:cos_sim_lu_score_llm_judge} illustrates the cosine similarity of VUFs from each layer of examples obtained with VU scores using different LLM-as-a-Judge models. We observe a high correlation between the three different scores for VUFs in the middle and subsequent layers. These results demonstrate that our observations are consistent regardless of the choice of verbal uncertainty score.

%%%%%%%%%%%%%%%%%%%%%%%%%%%%%%%%%%%%%%%%%%%%%%%%%%%%%
\subsection{Cosine Similarity between VUFs from different datasets.}
\label{app:lu_data_sim}
\begin{figure*}[ht]
    \centering
    \includegraphics[width=\textwidth]{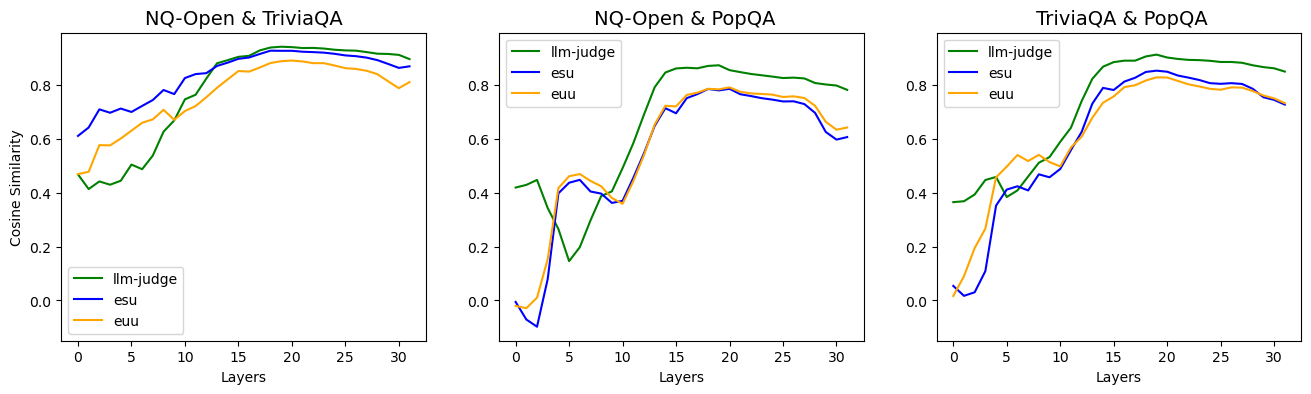}
    \caption{Cosine Similarity between VUFs from different datasets using different VU scores for Llama-3.1-8B-Instruct model.}
    \label{fig:cos_sim_lu_score_dataset}
\end{figure*}

To further support our observation that VUFs are consistent across datasete, we present cosine similarity between VUFs obtained from different datasets using different verbal uncertainty scores in Figure \ref{fig:cos_sim_lu_score_dataset}. We run experiments using Llama-3.1-8B-Instruct model. 

\subsection{Causal Validation with Alternative Methods of VUF Extraction}
\label{app:circularity}

It is in principle possible that the LLM that labels the samples used for determining the VUF and the LLM used to measure the VU score after the intervention actually measure some other consistent property of the text that is not VU. To exclude this possibility, we extracted VUF directions also using a very different method, based on measuring the mean cosine similarity with prototypical expressions of verbal uncertainty in a sentence embedding space obtained from an unrelated, encoder-only model introduced in Appendix~\ref{app:esu_euu}.

Figure~\ref{fig:causal_lu_ESU} presents the causal validation with the VUF extracted based on the ESU score instead of LLM-as-a-Judge. Similar to Figure~\ref{fig:causal_lu}, adding ESU-derived VUFs to model activations increases the VU (as judged by LLM) of the model outputs. Conversely, removing VUFs to activations decreases this uncertainty. These results are based on the Llama 3.1 8B model and the TriviaQA dataset.

\begin{figure}[ht]
    \centering
    \includegraphics[width=0.5\textwidth]{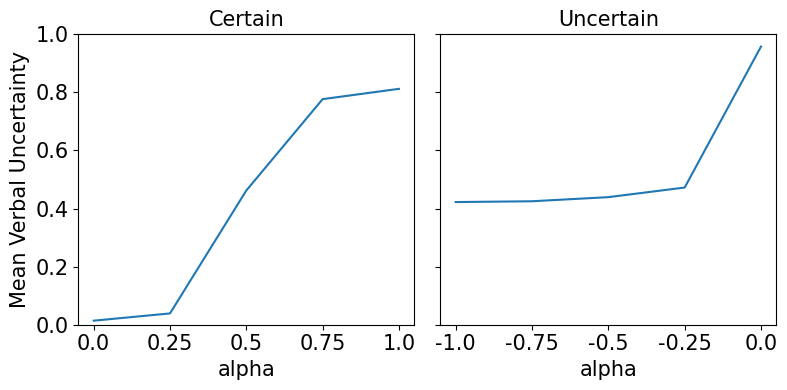}
    \caption{Mean model-generated answer verbal uncertainty on TriviaQA dataset with varying degrees of inference-time VUF intervention (modulated by the intervention intensity $\alpha$). The VUF is exacted via ESU.}
    \label{fig:causal_lu_ESU}
\end{figure}

\section{Hallucination Detection}
\subsection{Experimental Details for Probe Training}
\label{app:experiment_detail_detect}
These probes are linear models trained on the hidden states of LLMs to predict numerical uncertainty values in a single-run.
The hidden states are sourced from multiple layers within the LLM.
We have selected the following layers based on the performance for each uncertainty:
\begin{itemize}
\item  VU: Layers 5 to 20 for TriviaQA, 10 to 20 for NQ-Open, and 5 to 20 for PopQA.
\item  SU: Layers 10 to 20 for TriviaQA, 10 to 20 for NQ-Open, and 5 to 25 for PopQA.
\end{itemize}

For calculating metrics, we utilize the NumPy and NLTK packages.

\subsection{Classifier Binarized Uncertainty Probe}
\label{app:classifier_probe}

Given the hidden state, we train a logistic regression model (classifier probe) to predict binarized uncertainty. Instances with low verbal and high semantic uncertainty are labeled as hallucinations.

% \textbf{Classifier Binarized Uncertainty Probe}
As shown in Tab.~\ref{tab:SEP}, verbal uncertainty helps to improve the detection performance.
The ``Semantic only'' is the long-form setting of the SEP~\citep{kossen2024semantic} as the baseline. This work ignores the refusal cases and automatically classifies them into hallucinated which is not aligned with our definition.

\section{Hallucination Mitigation via Mechanistic Uncertainty Calibration (MUC)}
\subsection{Experimental Details for Mitigation}
\label{app: expert detail mitigate}
For the experiments in this work, we combine the VUFs exacted from three datasets together and construct $\mathcal{D}_{certain}$ and $\mathcal{D}_{uncertain}$ as samples with VU Score $\leq0.05$ and $\geq 0.9$ respectively.

We calculate semantic entropy (SE) to measure semantic uncertainty. The maximum of SE is the entropy of uniform distribution $max_SE = ln N$, is the number of samples used in the SE calculation. The minimum SE is 0. So $su(x)_{norm} = su(x)/ lnN$. The maximum of VU is 1.0, and the minimum is 0.
We set $max_\alpha=1.0$ for Llama-3.1-8B-Instruct, $max_\alpha=0.4$ for Mistral-7B-Instruct-v0.3, $max_\alpha=3.0$ for Qwen2.5-7B-Instruct, and $max_\alpha=4.0$ for Llama-3.1-70B-Instruct across three datasets.

\subsection{Case Study for Mitigation}
\label{app:mitigation case study}

In this section, we present examples of answers, highlighting the differences between the original generated responses and those after our mitigation method. 
Each example includes the question posed, the golden answer, the initial response generated by the LLM, and the response after  inference-time uncertainty calibration. 

% \paragraph{Example with middle VU}
\paragraph{Example with different calibration degrees (alphas)}
\begin{itemize}
  \setlength{\itemsep}{0pt} % Adjusts space between items
  \setlength{\parskip}{0pt} % Adjusts space between paragraphs
  \setlength{\parsep}{0pt}  % Adjusts space between paragraphs within an item

\item \textbf{Question:} Manhattan Island is bounded by three rivers, the Hudson, the East and which other?", 
\item \textbf{Golden Answer:} Harlem River
\item \textbf{Original Generated Response:} The other river is the Harlem River.
\item \textbf{Response after Calibration with alpha=0.14} I'm uncertain about the third river, but I can provide some information: Manhattan Island is bounded by three rivers, the Hudson and the East, with the Harlem River being the third river, however, it is sometimes considered a strait.
\item \textbf{Response after Calibration with alpha=0.70} I'm not certain about the third river, but I can provide some information: Manhattan Island is bounded by three rivers, the Hudson, the East, and the Harlem River.
\item \textbf{Response after Calibration with alpha=1.4} Unfortunately, I am unable to verify the name of the third river.
\end{itemize}

% \paragraph{Example with high VU}
% \begin{itemize}
%   \setlength{\itemsep}{0pt} % Adjusts space between items
%   \setlength{\parskip}{0pt} % Adjusts space between paragraphs
%   \setlength{\parsep}{0pt}  % Adjusts space between paragraphs within an item
% \item \textbf{Question:} The Bet365 Gold Cup, formerly the Beffred and the Whitbread Gold Cup, is the last major 'race of the National Hunt season. On which course is it run?
% \item \textbf{Golden Answer:} ['Sandown', 'Sandown, Isle of Wight']
% \item \textbf{Original Generated Response:} The Bet365 Gold Cup is run at Cheltenham.
% \item \textbf{Response after Calibration:} I am unable to verify the course on which the Bet365 Gold Cup is run.
% \end{itemize}

\subsection{Case Study for Decreased Correctness}
\label{app:decrease correctness}
\begin{itemize}

  \setlength{\itemsep}{0pt} % Adjusts space between items
  \setlength{\parskip}{0pt} % Adjusts space between paragraphs
  \setlength{\parsep}{0pt}  % Adjusts space between paragraphs within an item
\item \textbf{Question}: "Which horse won the 2013 English Grand National at Aintree?"

  \item \textbf{Golden Answer}: ["Aurora's Encore", "Auroras Encore"]

  \item \textbf{Semantic Entropy}: 1.83
  
  \item \textbf{Sampled Responses before Calibration}:

  \begin{itemize}
    \item "The horse that won the 2013 English Grand National at Aintree was named Auroras Encore. The Irish-bred and trained horse, \textbf{ridden by Daryl Jacob}, won the race with a time of 10 minutes and 3 seconds."
    
    \item "The horse that won the 2013 English Grand National at Aintree was named Auroras Encore. The horse, trained by Sue Smith, was \textbf{ridden by Ryan Mania}. Auroras Encore \textit{crossed the finish line with a winning margin of 5 lengths, making it a thrilling victory for the team}."
    
    \item "The horse that won the 2013 English Grand National at Aintree was Auroras Encore. The Irish-bred gelding was \textbf{ridden by jockey Daryl Jacob} and trained by Sue Smith. The win was a remarkable one as Auroras Encore \textit{was a 100/1 longshot, making it one of the biggest upsets in the history of the race}."
    
    \item "The horse that won the 2013 English Grand National at Aintree was named Auroras Encore. This Irish-bred thoroughbred, trained by Sue Smith, \textit{finished the grueling race with a time of 10 minutes and 5 seconds}. Auroras Encore was \textbf{ridden by jockey Ryan Mania}."
  \end{itemize}

  \item \textbf{Response after Calibration}: I'm afraid I don't have real-time data access, so I can't provide the exact winner of the 2013 English Grand National at Aintree...
  % However, the winner was either Auroras Encore or Neptune Collonges, as they crossed the finish line together and were jointly declared the winners.

\end{itemize}

\end{document}